\documentclass[sigconf]{acmart}
\usepackage{caption}
\usepackage{subcaption}
\usepackage{tikz}
\usepackage{subcaption}  % For subfigure environments and captions
\usepackage{xspace}
\usepackage{makecell}
\usepackage{multirow}
\usepackage{amsfonts}
\usepackage{subcaption}
\usepackage{url}
\usepackage[flushleft]{threeparttable}
\usepackage{supertabular}
\usepackage{placeins}
\usepackage{enumitem}
\usepackage{cleveref}
\usepackage{footnote}

\AtBeginDocument{%
  \providecommand\BibTeX{{%
    \normalfont B\kern-0.5em{\scshape i\kern-0.25em b}\kern-0.8em\TeX}}}
\setcopyright{none}

\acmConference{}{}{}{}{}{}
\acmBooktitle{}
\acmDOI{}
\acmISBN{}
\acmPrice{}
\acmYear{}
\copyrightyear{}
\newcommand{\method}{\texttt{KindSleep}\xspace}
\newcommand{\methodz}{\texttt{SLAM}\xspace}
\newcommand{\mlp}{\texttt{MLP-Regressor}\xspace}

\begin{document}

\title{
KindSleep: \underline{K}nowledge-\underline{In}formed \underline{D}iagnosis of Obstructive Sleep Apnea from Oximetry
}
\author{Micky C. Nnamdi}
\email{mnnamdi3@gatech.edu}
\affiliation{%
  \institution{Georgia Institute of Technology}
  \city{Atlanta}
  \state{Georgia}
  \country{USA}
}
\author{Wenqi Shi}
\email{wenqi.shi@utsouthwestern.edu}
\affiliation{%
  \institution{UT Southwestern Medical Center}
  \city{Dallas}
  \state{Texas}
  \country{USA}
}

\author{Cheng Wan}
\email{c.wan@gatech.edu}
\affiliation{%
  \institution{Georgia Institute of Technology}
  \city{Atlanta}
  \state{Georgia}
  \country{USA}
}

\author{J. Ben Tamo}
\email{jtamo3@gatech.edu}
\affiliation{%
  \institution{Georgia Institute of Technology}
  \city{Atlanta}
  \state{Georgia}
  \country{USA}
}

\author{Benjamin Smith}
\email{benjaminm.smith@shrinenet.org}
\affiliation{%
  \institution{Shriners Children’s}
  \city{Chicago}
  \state{Illinois}
  \country{USA}
}

\author{Chad Purnell}
\email{cpurnell@shrinenet.org}
\affiliation{%
  \institution{Shriners Children’s}
  \city{Chicago}
  \state{Illinois}
  \country{USA}
}

\author{May D. Wang}
\authornote{Corresponding author.}
\email{maywang@gatech.edu}
\affiliation{%
  \institution{Georgia Institute of Technology}
  \city{Atlanta}
  \state{Georgia}
  \country{USA}
}

\renewcommand{\shortauthors}{Nnamdi et al.}

\begin{abstract}
Obstructive sleep apnea (OSA) is a sleep disorder that affects nearly one billion people globally and significantly elevates cardiovascular risk. Traditional diagnosis through polysomnography is resource-intensive and limits widespread access, creating a critical need for accurate and efficient alternatives. 
In this paper, we introduce \method, a deep learning framework that integrates clinical knowledge with single-channel patient-specific oximetry signals and clinical data for precise OSA diagnosis. \method first learns to identify clinically interpretable concepts, such as desaturation indices and respiratory disturbance events, directly from raw oximetry signals. It then fuses these AI-derived concepts with multimodal clinical data to estimate the Apnea-Hypopnea Index (AHI).
We evaluate \method on three large, independent datasets from the National Sleep Research Resource (SHHS, CFS, MrOS; total n = 9,815).
\method demonstrates excellent performance in estimating AHI scores (R² = 0.917, ICC = 0.957) and consistently outperformed existing approaches in classifying OSA severity, achieving weighted F1-scores from 0.827 to 0.941 across diverse populations. 
By grounding its predictions in a layer of clinically meaningful concepts, \method provides a more transparent and trustworthy diagnostic tool for sleep medicine practices. 
\end{abstract}




\maketitle

\section{Introduction}
\label{sec:intro}
Obstructive sleep apnea (OSA) is a significant global health concern, affecting approximately one billion individuals worldwide between the ages of 30 and 69~\cite{lyons2020global}. OSA involves repeated episodes of partial or complete airway obstruction during sleep, increasing the risk of hypertension, cardiovascular disease, stroke, and significantly impairing overall quality of life~\cite{mitra2021association, zhao2017effect,shi2025predicting}. Therefore, accurate and timely diagnosis is essential for effective management and reduction of associated health complications.

\begin{figure*} 
    \centering
    \includegraphics[width=0.96\linewidth]{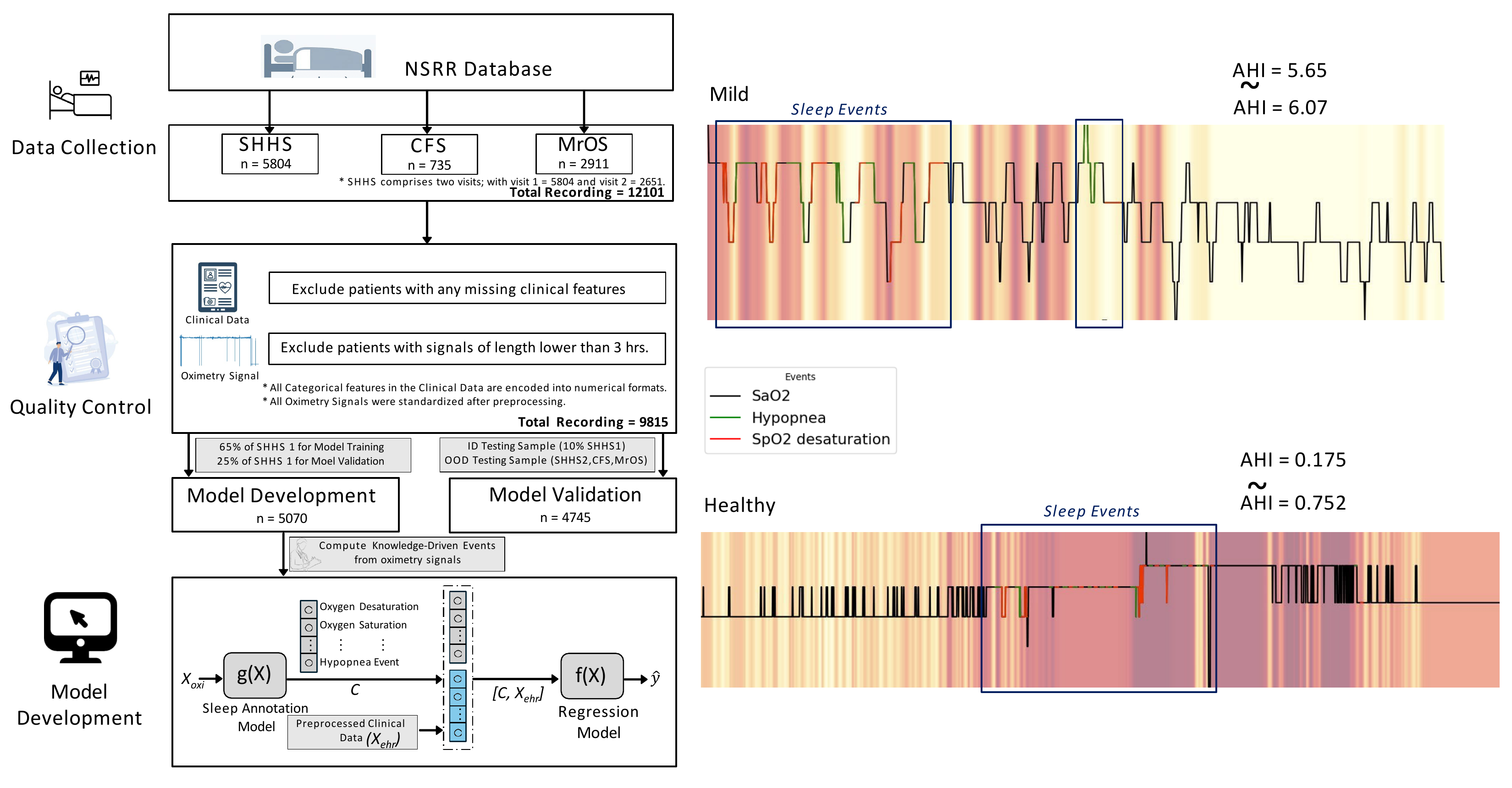}
        \caption{Overview of \method.  \method involved two main components: the sleep annotation model, which extracts clinically relevant metrics from raw oximetry signals, and the regression model, which integrates these metrics with processed clinical data to estimate the AHI. (Right) Example of oximetry signals from a mild OSA patient (top; reference AHI = 5.65) and a healthy control (bottom; reference AHI = 0.175), annotated with hypopnea events (green) and desaturations (red), alongside corresponding attention maps from the sleep annotation model that highlight the regions the model concentrates on, and the resulting AHI estimations from the regression model.
        }
     \label{fi1}
\end{figure*} 
Polysomnography (PSG), the gold standard for diagnosing OSA~\cite{brockmann2013diagnosis}, requires overnight monitoring of multiple physiological parameters, including electroencephalogram (EEG), electrocardiogram (ECG), electromyogram (EMG), electrooculogram (EOG), oronasal airflow, respiratory effort, and oxygen saturation levels. Each of these signals provides complementary information: EEG identifies sleep stages, airflow and oxygen saturation detect breathing disruptions, and EMG/EOG capture muscle tone and eye movements essential for event classification. Accurately computing the apnea hypopnea index (AHI) requires integrating these heterogeneous signals in a time-synchronized manner, a process that is technically demanding due to noise, inter-signal variability, and the need for expert scoring. Clinicians rely on this integrated signal analysis measurements to compute the AHI, classifying OSA severity into normal (AHI $<$ 5), mild (5 $\leq$ AHI $<$ 15), moderate (15 $\leq$ AHI $<$ 30), and severe (AHI $\geq$ 30)~\cite{hudgel2016sleep}. However, PSG is complex, resource-intensive, and expensive, severely restricting its accessibility, especially in resource-limited regions.

Given these limitations, researchers have increasingly focused on automated diagnostic methods leveraging fewer physiological channels, notably oximetry signals~\cite{chen2023deep, gutierrez2021ensemble, alarcon2023obstructive, barnes2022detection, li2022sleep, wan2024advancing,zan2025weakly,van2018automated}. Recently, multimodal approaches combining oximetry signals with clinical data have emerged, aiming to enhance diagnostic accuracy by incorporating comprehensive patient-specific data~\cite{levy2023deep,fayyaz2024multimodal}. Oximetry provides crucial insights into respiratory disturbances during sleep through measurements of blood oxygen saturation fluctuations~\cite{nardini2023improving}. Concurrently, clinical data offer valuable demographic, clinical, and lifestyle information, including critical risk factors and comorbidities linked to OSA~\cite{keenan2020multisite, te2024identifying}. Body mass index (BMI), in particular, is strongly correlated with OSA prevalence, significantly higher among overweight and obese populations~\cite{ahima2013health, uzair2024correlation, young2002epidemiology, senaratna2017prevalence, ming2021metabolic, fattal2022body}. Such findings highlight the importance of integrating clinical and demographic data into OSA diagnostic models.
However, existing automated OSA diagnostic methods using oximetry often face challenges, including limited accuracy, generalizability, and transparency in decision-making. Such "black-box" AI approaches hinder clinical trust, preventing clinicians from understanding, validating, and confidently using these tools in clinical practice~\cite{uddin2018classification,xu2025medagentgym,giuste2022explainable,shi2021learning,nnamdi2023model,tamo2023uncertainty,vyshnya2024optimized,shi2024ehragent}.

To address these issues, we introduce \method, an AI-based framework designed to integrate clinical expertise with oximetry and clinical data for accurate and reliable OSA diagnosis. \method employs the concept bottleneck model (CBM) paradigm~\cite{koh2020concept,nnamdi2023concept}, leveraging clinically interpretable metrics derived from oximetry data, such as oxygen desaturation indices, oxygen saturation levels, and hypopnea event frequencies, as intermediate features to guide model predictions~\cite{giuste2021automated}.
In this study, we validated \method using extensive data from three independent cohorts (SHHS, CFS, MrOS) encompassing $9,815$ sleep recordings from the National Sleep Research Resource (NSRR). \method demonstrated strong predictive performance, achieving a coefficient of determination ($R^2$) of $0.917$ and an intraclass correlation coefficient (ICC) of $0.957$, significantly outperforming existing methods. Notably, our framework exhibited high generalizability across diverse patient populations, achieving weighted F1-scores ranging from 0.827 to 0.941.

\section{\method}
\label{sec:method}
\subsection{Overview}
We propose \method, a clinically grounded machine learning framework designed to estimate AHI using a combination of oximetry signals and clinical data. The key innovation lies in introducing an intermediate layer of clinically interpretable features—knowledge-informed metrics—derived from the raw oximetry signal. These metrics simulate the annotations a sleep technologist would typically provide and help bridge the gap between raw physiological input and meaningful clinical prediction.  \method comprises two key components: a SLeep Annotation Model (\methodz) that derives interpretable concepts and a regressor model (primarily implemented as \mlp) that leverages these concepts—together with clinical data—to produce the final AHI prediction.
\begin{table}
\centering
\caption{Descriptions of Extracted Signal Annotations \protect\footnotemark.}
\label{table4}
\begin{tabular}{p{1cm} p{7cm}}
\hline\hline
\textbf{Label} &  \textbf{Description}  \\ \hline
ahi\_a0h4 & (Apneas with no oxygen desaturation threshold used and with or without arousal $+$ hypopneas with $>$ 30\% flow reduction and $\geq$ 4\% oxygen desaturation and with or without arousal) $/$ hour of sleep \\ \hline
ahi\_a0h4a & Similar to ahi\_a0h4, but hypopneas are tallied if they meet either of two conditions: $\geq$ 4\% oxygen desaturation or evidence of arousal. \\ \hline
ahi\_c0h3 & (Central apneas with no oxygen desaturation threshold used and with or without arousal $+$ hypopneas with $>$ 30\% flow reduction and $\geq$ 3\% oxygen desaturation and with or without arousal) / hour of sleep \\ \hline
ahi\_c0h4 & Central apnea index under the same conditions as above, except that hypopneas are credited if they meet a $\geq$ 4\% desaturation threshold or are associated with arousal. \\ \hline
avgsat & Mean oxygen saturation value calculated over the full sleep period  \\ \hline
minsat & Minimum oxygen saturation value recorded during the sleep period
\\ \hline
rdi0p & Apneas with no oxygen desaturation threshold used and with or without arousal $+$ hypopneas with $>$ 30\% flow reduction and with no oxygen desaturation used and with or without arousal / hour of sleep
\\ \hline
rdi2p & Apneas with $\geq$ 2\% oxygen desaturation and with or without arousal $+$ hypopneas with $>$ 30\% flow reduction and $\geq$ 2\% oxygen desaturation and with or without arousal / hour of sleep 
\\ \hline
rdi3p & Same as above, but requiring $\geq$ 3\% oxygen desaturation
\\ \hline
rdi4p & Same as above, but requiring $\geq$ 4\% oxygen desaturation
\\ \hline \hline
\end{tabular}
\end{table}
\footnotetext{\url{https://sleepdata.org/datasets/shhs/variables}}
Given an oximetry signal $X_{oxi}\in \mathbb{R}^{1 \times d}$, representing a single-channel oxygen saturation ($\text{SpO}_{2}$) measurement recorded at a $1\text{Hz}$ sampling rate, where $d$ is the length of the recording in seconds ($25200$), and a set of knowledge-informed metrics $C \in \mathbb{R}^{m}$, which represent sleep relevant features computed from sleep events such as the oxygen desaturation index, average oxygen saturation, minimum oxygen saturation, and the frequency of apnea and hypopnea events. Our first task is to learn a function that estimates this set of knowledge-informed metrics $C$ from the raw signals $X_{oxi}$ using \methodz $g$, such that:
\begin{equation}
    g: \mathbb{R}^{1 \times d} \rightarrow \mathbb{R}^{m},
\end{equation}
such that $C = g(X_{oxi})$. The model $g$ is trained by minimizing the loss between the estimated and reference knowledge-informed metrics:
\begin{equation}
    \min_{g \in \mathcal{G}} \frac{1}{N} \sum_{i=1}^{N} \mathcal{L}_C(C_i, g(X_{oxi}^{(i)})),
\end{equation}
where $\mathcal{L}_C$ is an appropriate loss function, $C_i$ represents the ground truth metrics for subject $i$, and $\mathcal{G}$ denotes the space of \methodz $g$.

The predicted knowledge-informed metrics $C$, along with patient clinical data $X_{ehr} \in \mathbb{R}^{e}$, which includes demographic and clinical information such as age, gender, BMI, and comorbidities are concatenated into a single feature vector $[C, X_{ehr}] \in \mathbb{R}^{m+e}$. This combined representation is used to estimate the AHI, a continuous measure of OSA severity. The regression model is defined as:
\begin{equation}
    f: \mathbb{R}^{m+e} \rightarrow \mathbb{R},
\end{equation}
where $f$ estimate the AHI score $\hat{y}$ from the concatenated input $[C, X_{ehr}]$. The objective is to minimize the prediction error over the dataset:
\begin{equation}
    \min_{f \in \mathcal{F}} \frac{1}{N} \sum_{i=1}^{N} \mathcal{L}_Y\left(y_i, f([g(X_{oxi}^{(i)}), X_{ehr}^{(i)}])\right),
\end{equation}
where $N$ is the number of subjects, $y_i$ is the ground-truth AHI for subject $i$, $g(X_{oxi}^{(i)})$ is the predicted knowledge-informed metric vector, $X_{ehr}^{(i)}$ is the corresponding clinical data, $\mathcal{L}_Y$ is a loss function, and $\mathcal{F}$ denotes the space of regression model $f$.

Subsequently, the predicted AHI score $\hat{y}$ is discretized into clinically defined categories representing OSA severity levels (e.g., no apnea, mild, moderate, severe), using standard thresholding criteria~\cite{hudgel2016sleep}. It is important to note that during training, the ground-truth values of the knowledge-informed metrics $C$ are available and are used to supervise the learning of the annotation model $g$. However, during deployment, these annotations are not assumed to be accessible. Instead, the model must infer them directly from the raw oximetry signal $X_{oxi}$. This setup ensures that the full pipeline beginning with signal input and ending with AHI prediction, remains fully automated and aligned with real-world clinical deployment conditions.

\subsection{Knowledge-Informed Metrics}
Knowledge-informed metrics refer to clinically meaningful annotations derived from raw physiological signals—specifically, oximetry data that capture essential indicators of sleep-disordered breathing. These metrics are analogous to the assessments a trained sleep technologist would extract during a manual scoring process, and they represent quantifiable components of sleep apnea severity. By isolating these intermediate features, our model introduces a level of interpretability and clinical alignment rather than relying solely on end-to-end black-box learning from raw signals.

The use of these metrics serves two core purposes. First, they provide a clinically interpretable layer that enables insight into the physiological events driving the model’s predictions. Second, they act as a bottleneck mechanism that encourages the model to learn meaningful representations aligned with medical understanding of sleep apnea, particularly the AHI, which is the standard diagnostic metric for OSA severity.

These metrics include variations of AHI and RDI (Respiratory Disturbance Index) computed under different desaturation thresholds and event definitions, as well as statistical features such as average and minimum oxygen saturation during sleep. All metrics are derived from temporal patterns of oxygen desaturation, apnea, and hypopnea events. Table~\ref{table4} provides a detailed breakdown of each extracted annotation and its clinical interpretation.

\begin{figure*}[h]
    \centering
    % (a) Parity plots
    \begin{subfigure}[b]{\textwidth}
        \centering
        \includegraphics[width=0.24\linewidth]{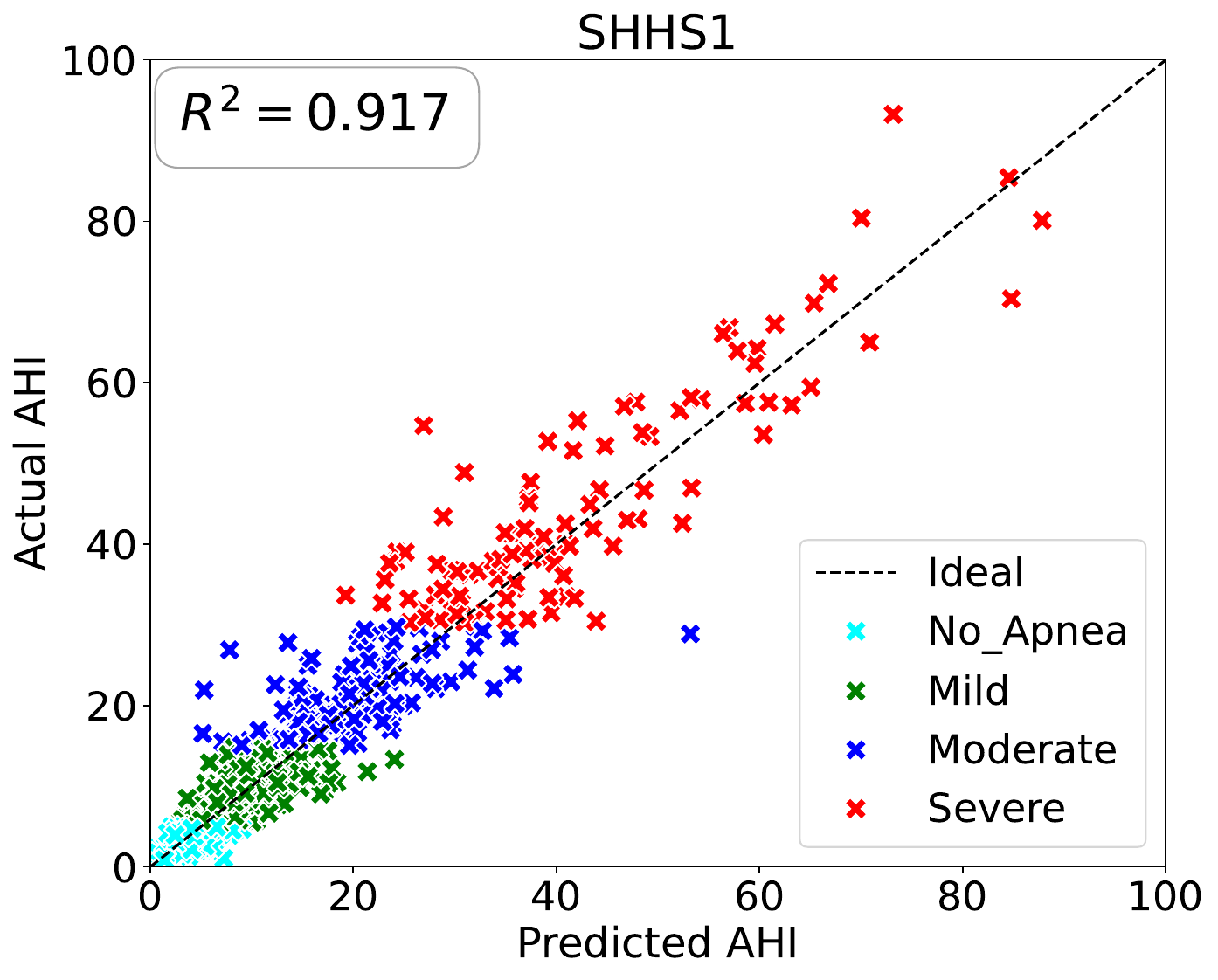}
        \hfill
        \includegraphics[width=0.24\linewidth]{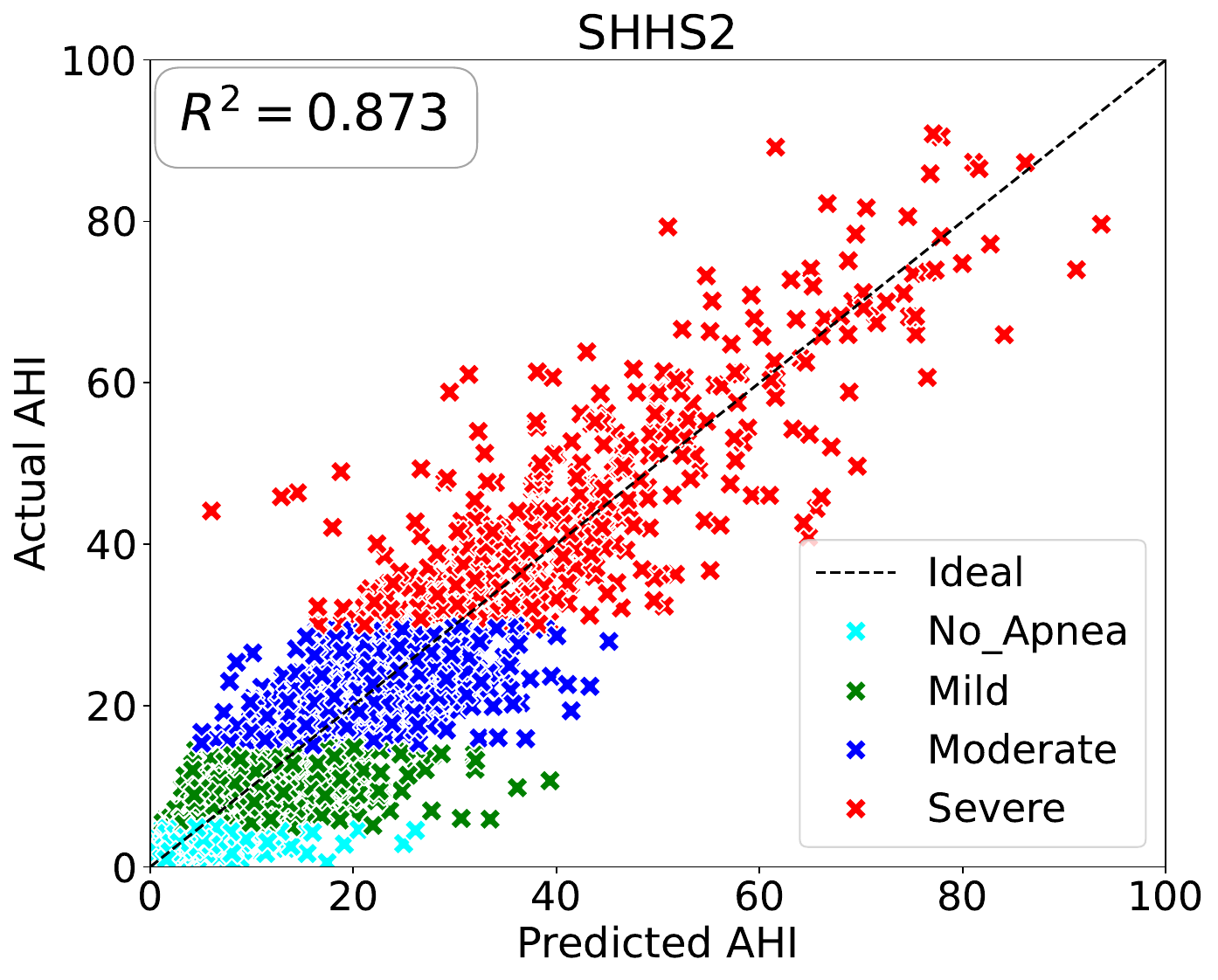}
        \hfill
        \includegraphics[width=0.24\linewidth]{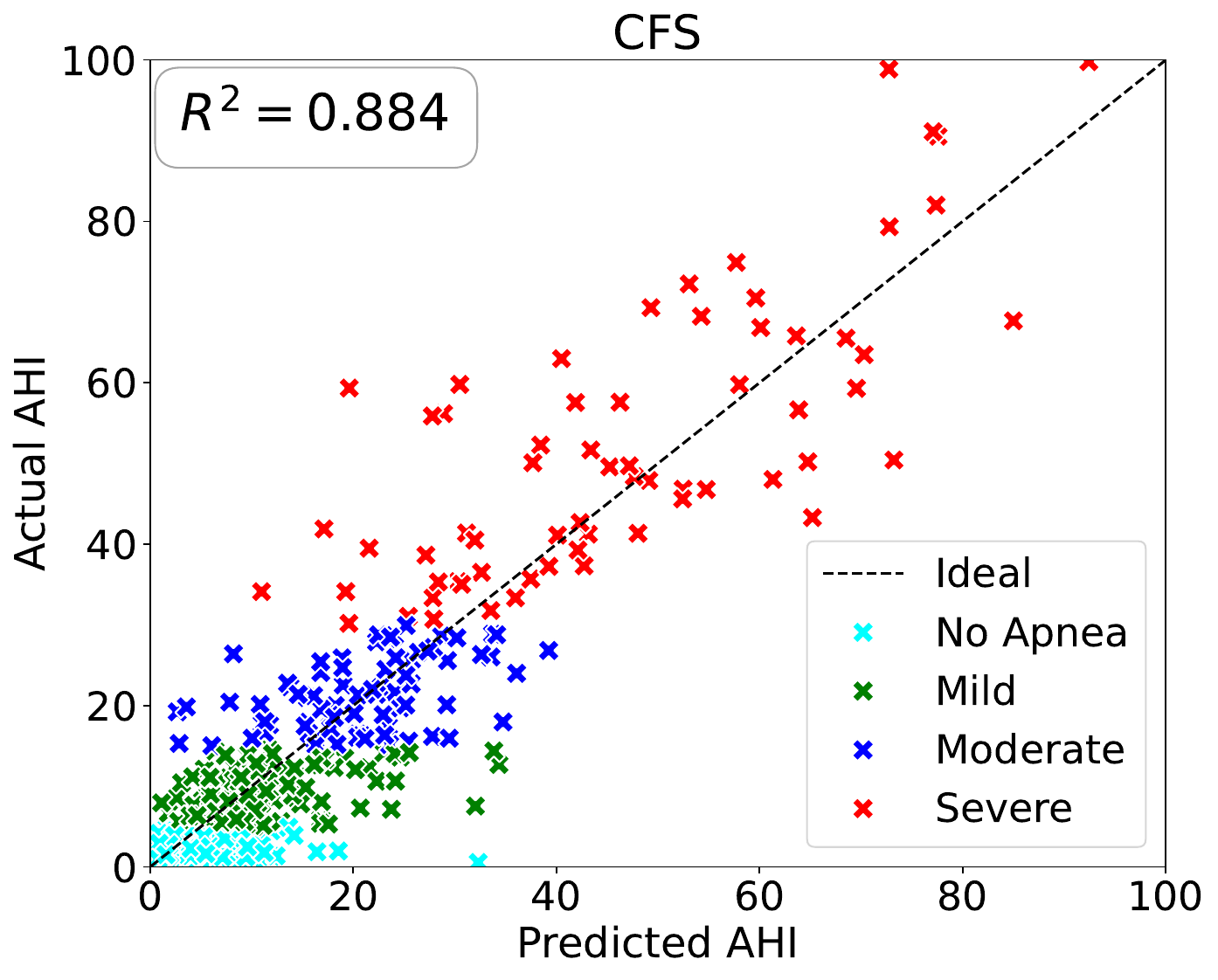}
        \hfill
        \includegraphics[width=0.24\linewidth]{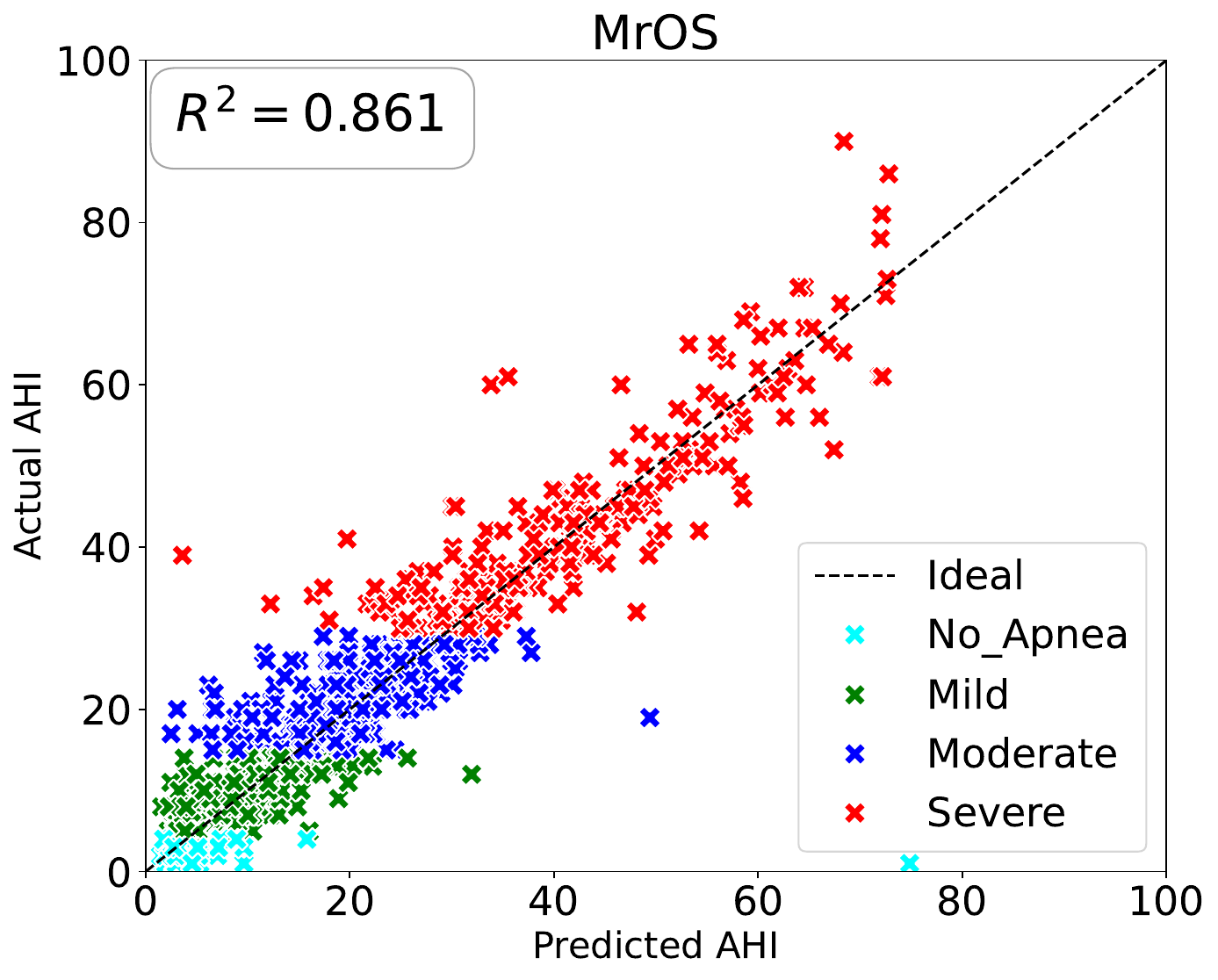}
        \captionsetup{labelformat=empty}
        \caption{(a) \mlp: The scatter plots show the relationship between the predicted AHI and the actual AHI for the SHHS1, SHHS2, CFS, and MrOS datasets. The data points are color-coded based on the OSA severity categories. The black dashed line represents the ideal prediction scenario where the predicted AHI perfectly aligns with the actual AHI.}
    \end{subfigure}
    
    \vspace{1em}
    
    % (b) Bland–Altman plots
    \begin{subfigure}[b]{\textwidth}
        \centering
        \includegraphics[width=0.24\linewidth]{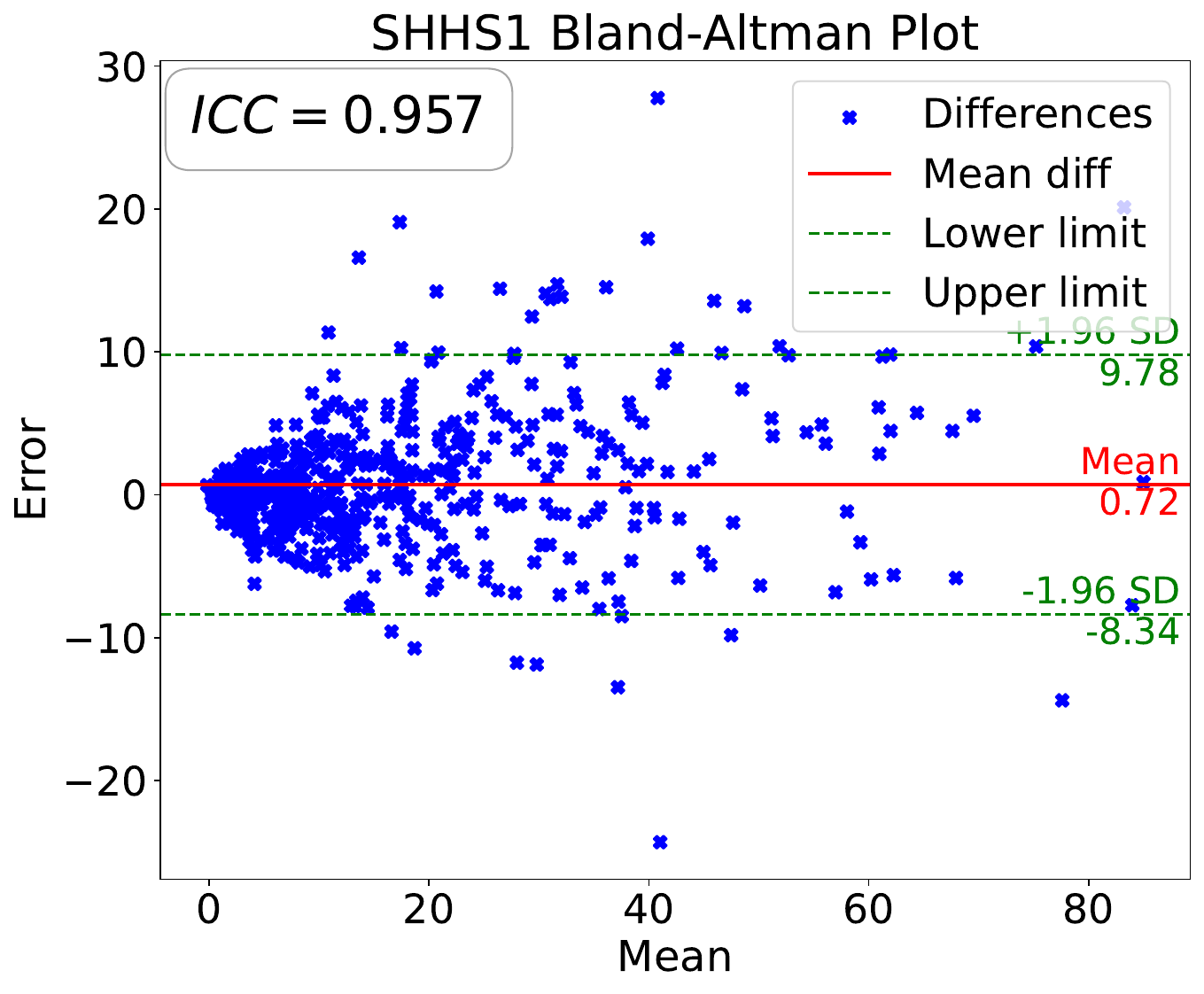}
        \hfill
        \includegraphics[width=0.24\linewidth]{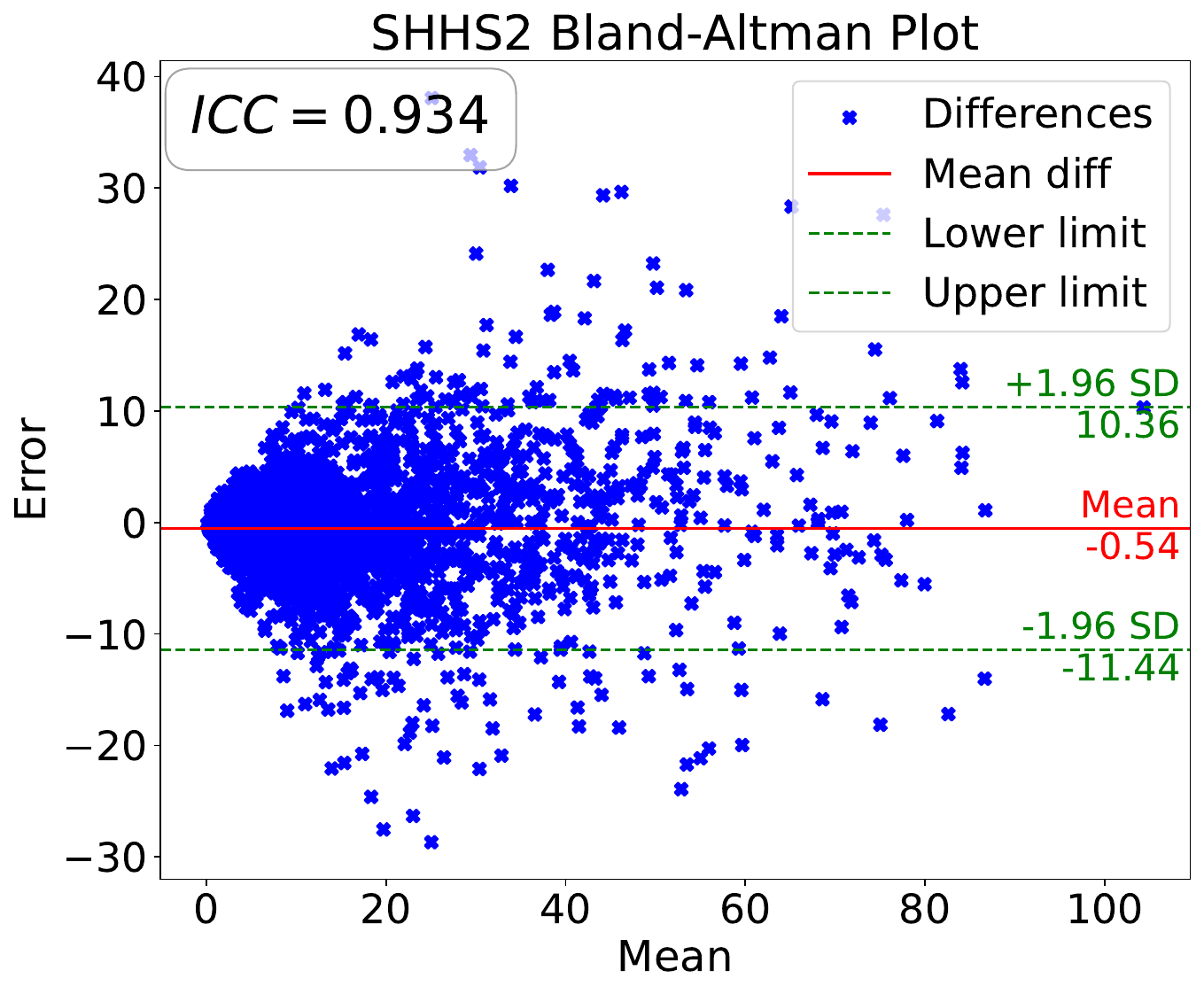}
        \hfill
        \includegraphics[width=0.24\linewidth]{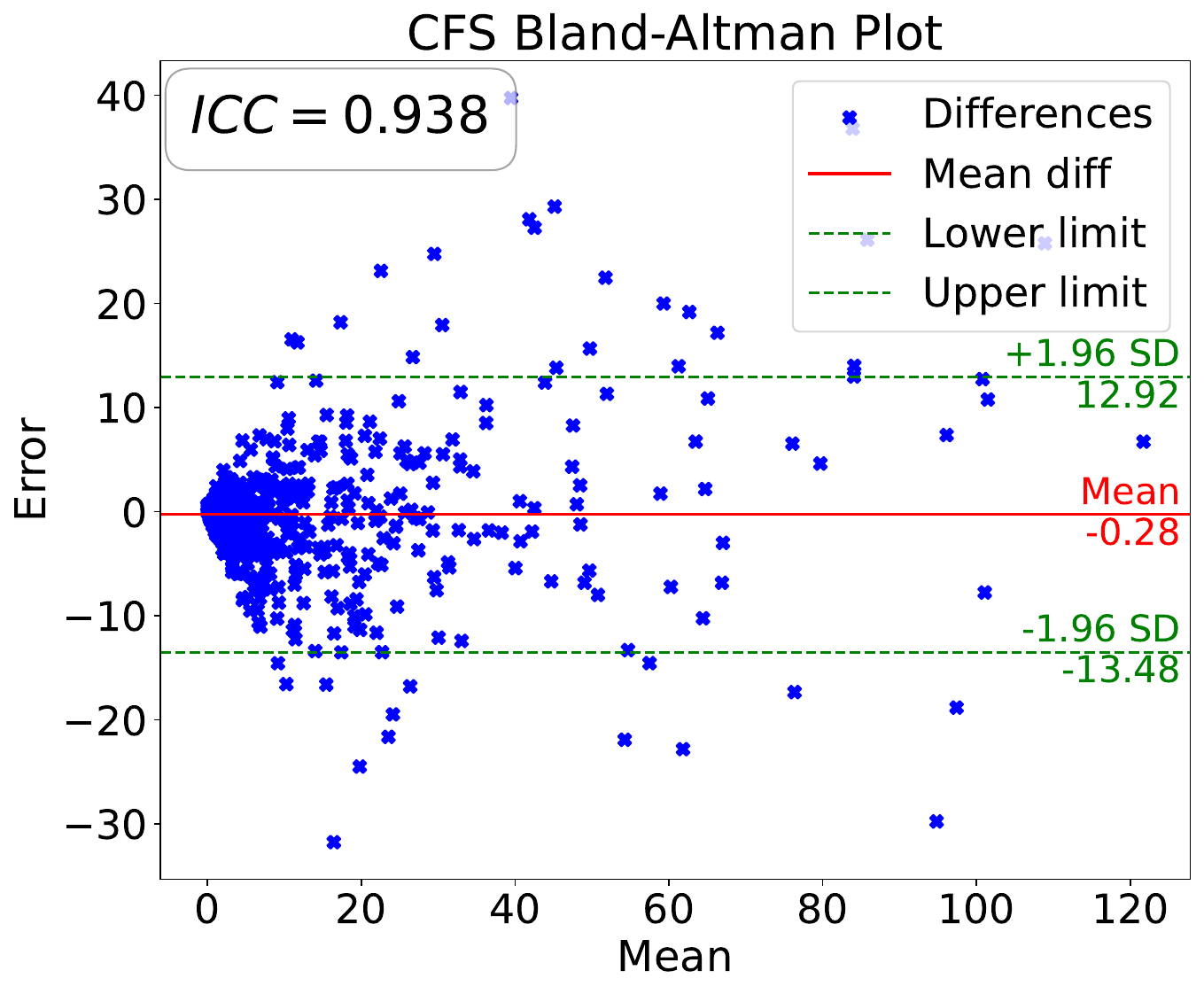}
        \hfill
        \includegraphics[width=0.24\linewidth]{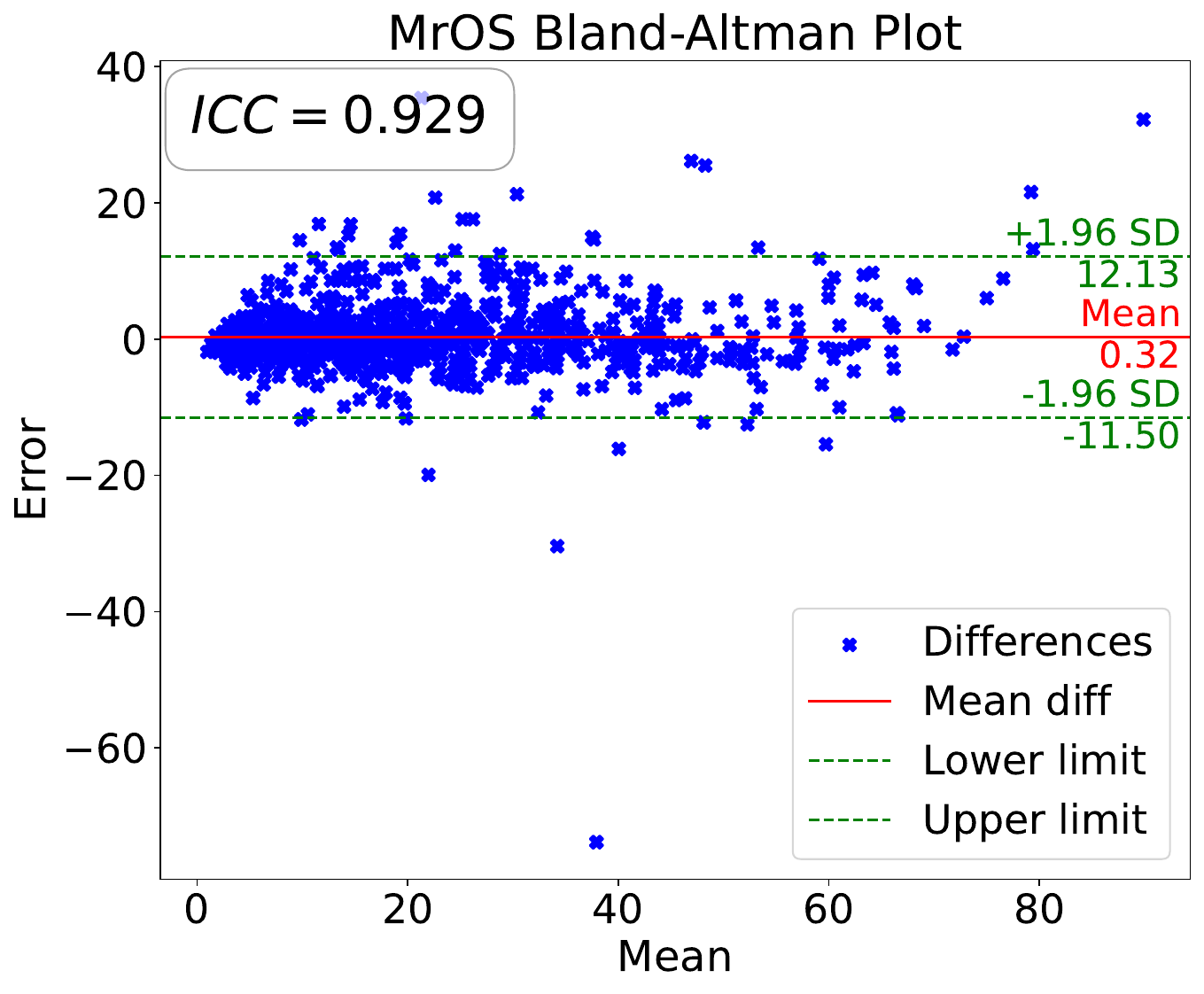}
        \captionsetup{labelformat=empty}
        \caption{(b) \mlp: The Bland-Altman plots compare the original AHI measurements with the estimated AHI measurements for SHHS1, SHHS2, CFS, and MrOS. The plot displays the difference in AHI error between the original and estimated measurements against their mean AHI. The dashed green line represents the mean difference, indicating the systematic bias. The solid red lines show the limits of agreement ($\pm 1.96$ standard deviations from the mean difference), which define the range within which most differences between the methods are expected to lie. Each blue dot represents an individual measurement pair, illustrating the agreement and variability between the original and estimated AHI values.}
    \end{subfigure}
    
    \vspace{1em}
    
    % (c) Confusion matrices
    \begin{subfigure}[b]{\textwidth}
        \centering
        \includegraphics[width=0.24\linewidth]{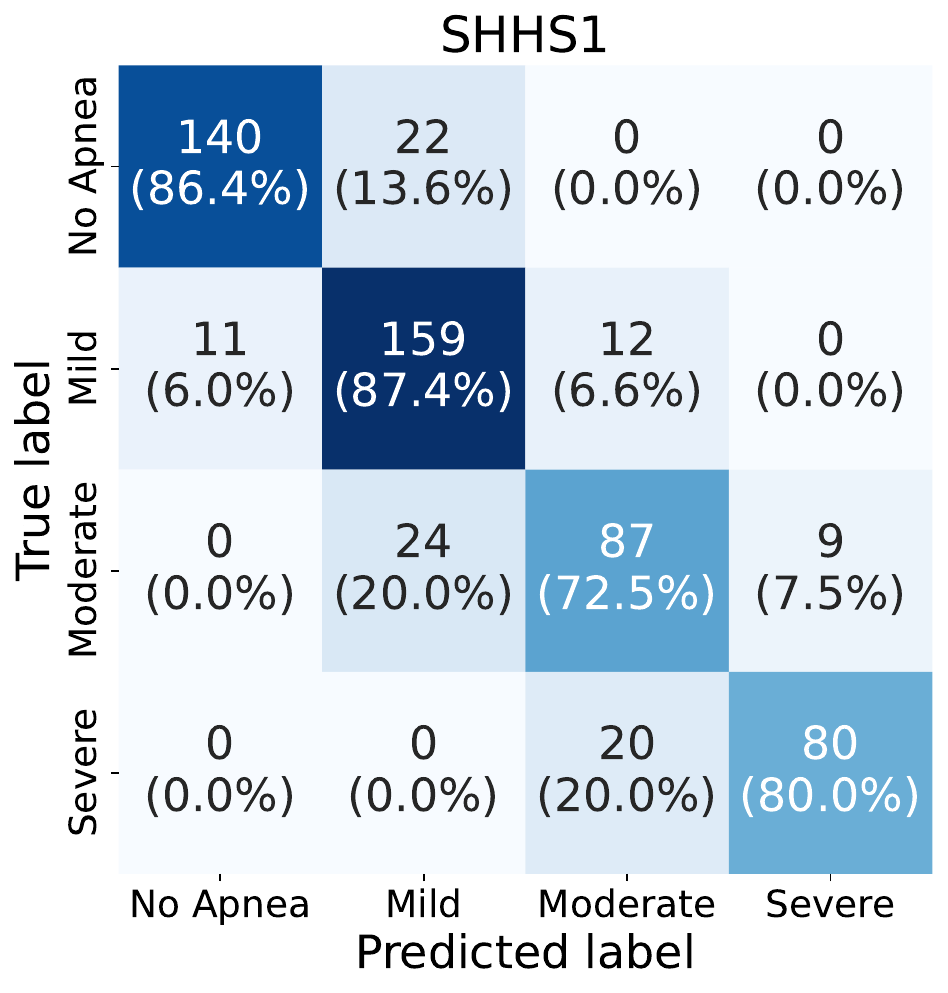}
        \hfill
        \includegraphics[width=0.24\linewidth]{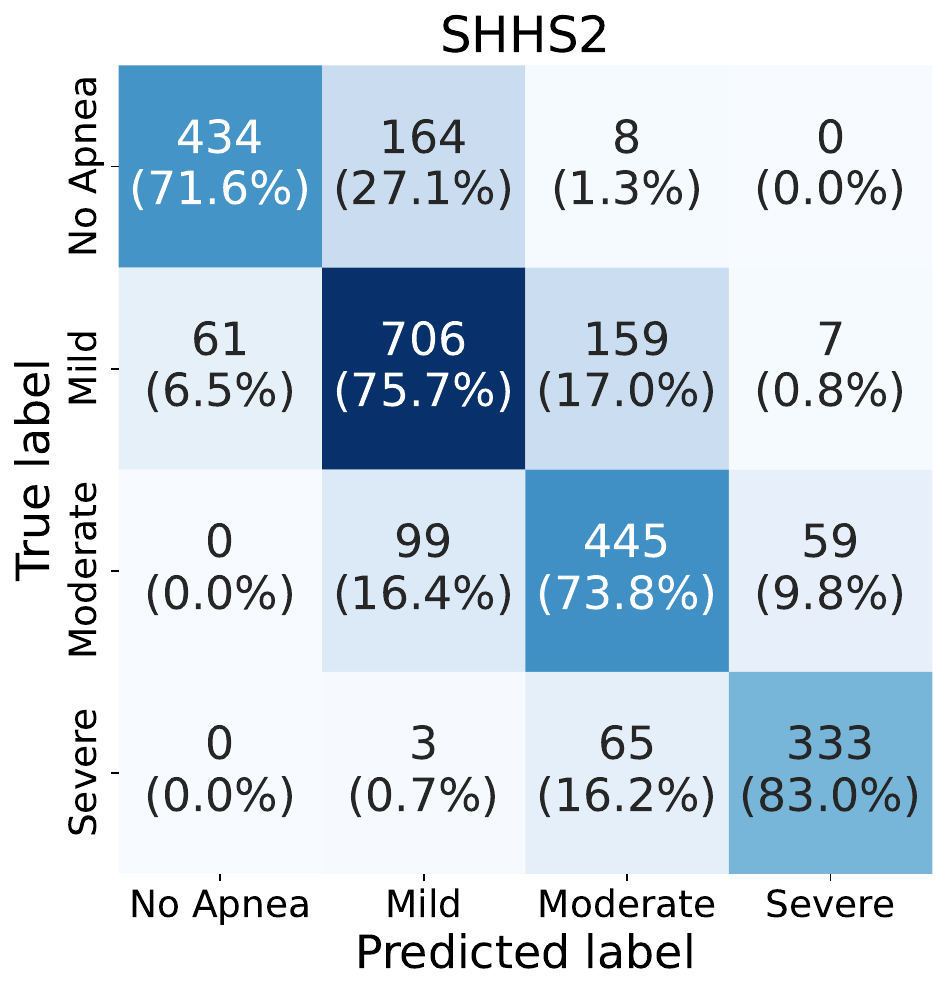}
        \hfill
        \includegraphics[width=0.24\linewidth]{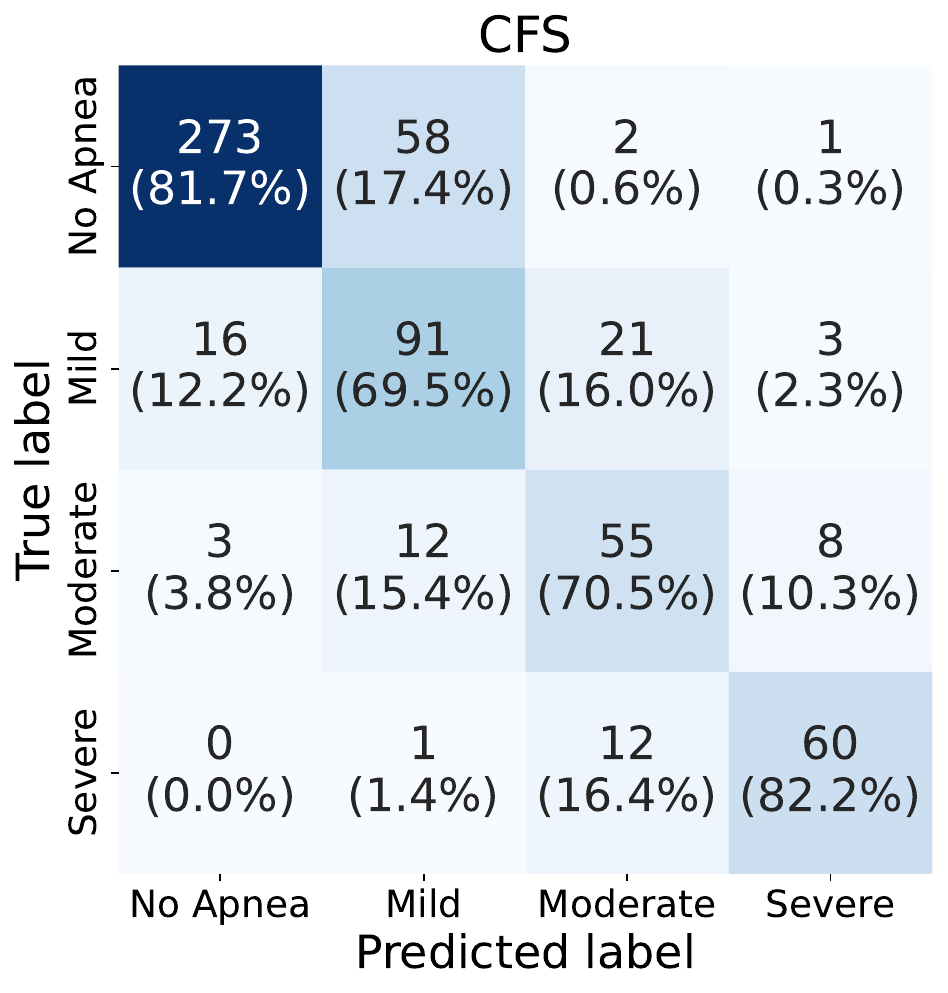}
        \hfill
        \includegraphics[width=0.24\linewidth]{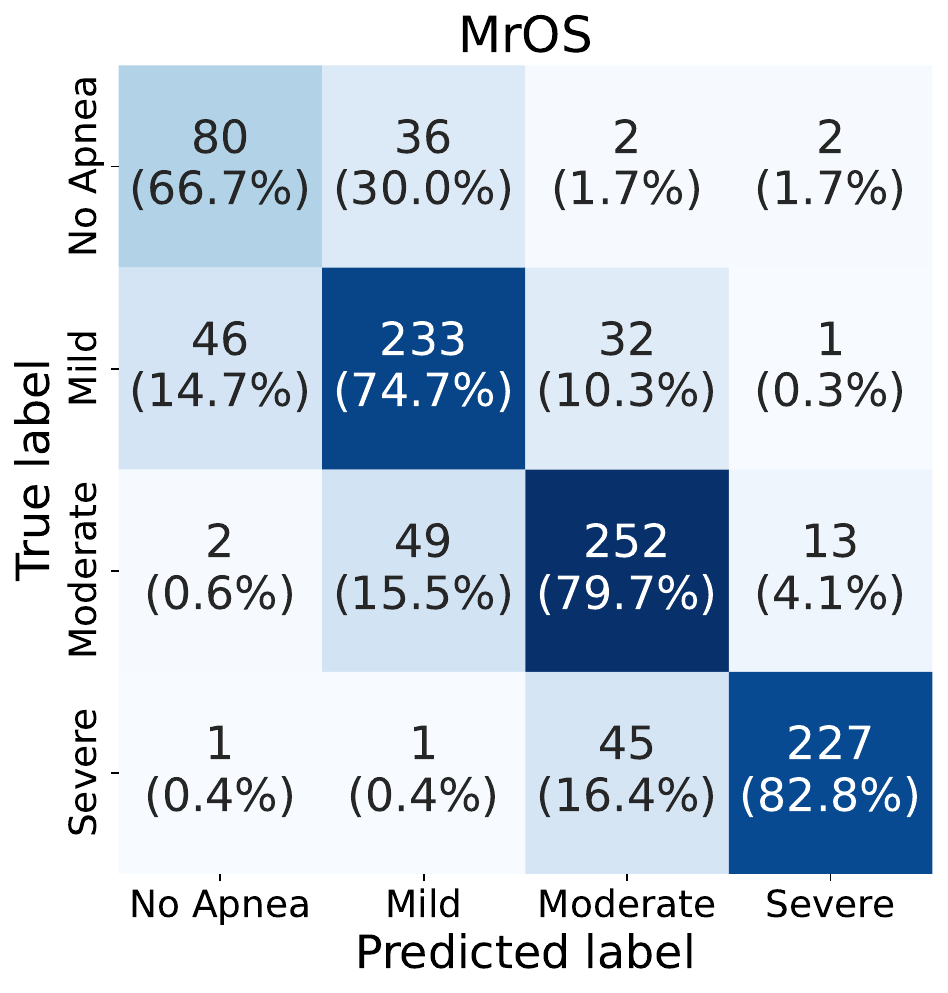}
        \captionsetup{labelformat=empty}
        \caption{(c) \mlp: The confusion matrix results of our \method framework were evaluated using three different testing datasets (SHHS 1, SHHS 2, CFS, and MrOS) to assess generalization. The results demonstrate varying classification outcomes across the four predicted classes.}
    \end{subfigure}
    
    \caption{(a) Parity plots, (b) Bland–Altman plots, and (c) confusion matrix results for SHHS1, SHHS2, CFS and MrOS.}
    \label{fi6}
\end{figure*}

\begin{table}
\centering
\caption{Performance Metrics of \methodz Across Different Datasets. The table shows the $MAE$ and $RMSE$ for SHHS1, SHHS2, CFS, and MrOS, indicating the robustness and generalization capability of \methodz, when compared with other models.}
\resizebox{1.01\linewidth}{!}{ %
\begin{tabular}{lcccccccc}
\hline\hline
\textbf{Models} & \multicolumn{2}{c}{\textbf{SHHS1}} & \multicolumn{2}{c}{\textbf{SHHS2}} & \multicolumn{2}{c}{\textbf{CFS}} & \multicolumn{2}{c}{\textbf{MrOS}} \\ \cline{2-9}
                & \textbf{MAE}   & \textbf{RMSE}  & \textbf{MAE}   & \textbf{RMSE}  & \textbf{MAE }  & \textbf{RMSE}  & \textbf{MAE}   & \textbf{RMSE}  \\ \hline
Linear   & 21.526 & 29.341 & 24.140 & 36.512 & 33.319 & 45.180 & 20.320 & 51.314\\ 
Ridge    & 15.741 & 21.198 & 15.204 & 20.767 & 21.014 & 28.612 & 5.116  & 11.509\\
Decision Tree       & 9.740 & 14.177 & 10.049 & 14.540 & 11.236 & 17.108 & 3.737  & 8.729\\
K-Neighbors  & 9.238 & 13.805 & 8.793 & 13.238 & 10.363 & 15.179 & 8.161  & 11.095 \\
SVR & 8.425 & 12.710 & 8.095 & 12.165 & 9.744 & 14.946 & 8.401  & 12.078\\
GradBoost  & 7.808 & 10.905 & 7.934 & 11.032 & 9.634 & 13.358 & 7.266  & 9.249 \\
XGBoost  & 7.335 & 10.374 & 7.327 & 10.405 & 9.155 & 13.173 & 3.366  & 6.232 \\
CatBoost & 7.302 & 10.306 & 7.456 & 10.367 & 9.296 & 12.894 & 6.483  & 8.407\\
LightGBM & 6.782 & 9.617 & 6.843 & 9.663 & 8.820 & 12.550 & 3.866  & 6.169\\
\methodz   & \textbf{3.500} & \textbf{5.269}  & \textbf{3.787} & \textbf{5.555} & \textbf{4.429} & \textbf{7.495}  & \textbf{3.264} & \textbf{5.485 }   \\ 
 \hline\hline
\end{tabular}
}
\label{ap_3}
\end{table}

\begin{table*}[]
\centering
\caption{Evaluation Metrics Across Various Datasets: The weighted $F_1$ Score, precision, sensitivity, specificity, $R^2$,  $ICC$, $MAE$ and $RMSE$ across various datasets for the regression model considered. All metrics are weighted, meaning they are calculated by averaging values proportionally to class sizes to account for class imbalances in the datasets.}
\resizebox{\linewidth}{!}{
\begin{tabular}{cccccccccc}
\hline\hline
\textbf{Regression Model} & \multicolumn{1}{c}{\textbf{Dataset}} & \textbf{F1 Score} & \textbf{Precision} & \textbf{Sensitivity} & \textbf{Specificity} & \textbf{$R^2$} & \textbf{ICC} &\ \textbf{MAE} &\ \textbf{RMSE}  \\ \hline
\multirow{4}{*}{\mlp} & SHHS1 & $0.827 \pm 0.030$    & $0.832 \pm 0.030$     & $0.826 \pm 0.034$ & $0.935 \pm 0.013$ & 0.917 & 0.957 & 3.101 & 4.677\\

& SHHS2 & $0.756 \pm 0.016$    & $0.763 \pm 0.017$    & $0.754 \pm 0.017$      & $0.899 \pm 0.008$ & $0.873$ & 0.934 & 3.731 & 5.589\\ 

& CFS & $0.786 \pm 0.032$     & $0.803 \pm 0.031$   & $0.778 \pm 0.034$      & $0.921 \pm 0.017$ & $0.884$ & 0.938 & 3.930 & 6.743\\ 

& MrOS & $0.777 \pm 0.026$ &  $0.782 \pm 0.025$    & $0.775 \pm 0.025$      & $0.916 \pm 0.010$ & 0.861 & 0.929 & 3.486 & 6.038\\  
\hline \hline
\end{tabular}}
\label{1}
\end{table*}

\section{Results}
\subsection{Datasets}
In this study, we used three large-scale, independent cohorts from NSRR \cite{zhang2018national}, including:
\begin{itemize}
\item \textbf{Sleep Heart Health Study (SHHS)} \cite{quan1997sleep}, organized through the National Heart, Lung, and Blood Institute, was designed to investigate how sleep-disordered breathing affects cardiovascular health and related conditions. The cohort originally enrolled 6,441 adults aged 40 years and above for its first phase, during which overnight oximetry measurements were gathered. A second phase followed with 3,295 participants. Although SHHS collected comprehensive overnight PSG data, the oximetry signal analyzed in this work was measured using the Nonin XPOD Model 3011 device.

\item \textbf{Cleveland Family Study (CFS)} \cite{redline1995familial} consists of data collected across as many as four study visits spanning 16 years, including 2,284 participants drawn from 361 families, with nearly half of the participants identifying as African American. Oximetry signal was measured using a Nonin 8000 sensor at a sampling frequency of 1 Hz. For the present analysis, we relied on a curated subset released through NSRR, which consists of 735 overnight oximetry recordings.

\item \textbf{Osteoporotic Fractures in Men Study (MrOS)} \cite{blackwell2011associations} is an ancillary study of the primary Osteoporotic Fractures in Men Study. During the initial recruitment period (2000–2002), 5,994 men aged 65 years or older, all living independently in the community, were enrolled across six clinical sites for baseline assessments. A few years later, from late 2003 through early 2005, 3,135 of these participants took part in the ancillary Sleep Study, which included unattended overnight PSG along with three to five days of wrist actigraphy monitoring.
\end{itemize}

\subsection{Experimental Setup}
\subsubsection{Dataset Split:}  \method was trained using 65\% from the SHHS1 dataset as the training set and 25\% as the validation set. The remaining 10\% of SHHS1 served as the in-distribution test set. Additionally, data from the SHHS2, CFS, and MrOS datasets were used as the out-of-distribution test sets to evaluate the model’s generalization across diverse populations. 
% Further details on the data split process are provided in the \cref{sec:data}. 

\subsubsection{Implementation Details}

To optimize the model performance, we employed a Bayesian hyperparameter optimization strategy for both \methodz and \mlp. The search space included model specific architectural parameters (e.g., number of layers, kernel sizes, number of filters), as well as training hyperparameters (e.g., learning rates, dropout rates, regularization strength, and optimizer type). Each configuration was evaluated using the validation set, and the model achieving the lowest validation loss was selected for final testing. 
All experiments were conducted on an NVIDIA A100 GPU with 80GB of memory, and random seeds were fixed to 42 to ensure reproducibility.

\subsubsection{Parameter Settings}
For the \methodz, hyperparameter tuning was performed using Optuna. The search space included variations in the kernel sizes, number of filters, dropout rates, learning rates, weight decay, and activation functions. The best configuration used  256 filters, a kernel size of 9, a dropout rate of 0.1, and a learning rate of 0.0005 without weight decay, using the Adam optimizer and ReLU activation.
For the \mlp, the optimal configuration included ReLU activation, an L2 penalty of 0.1, a single hidden layer with 50 units, the lbfgs solver, and an adaptive learning rate of 0.001.

\subsection{Main Results}
\subsubsection{Effect of \methodz}
We evaluate the robustness and generalization capability of \methodz across various datasets. As shown in Table~\ref{ap_3}, \methodz consistently performs well across both in-distribution and out-of-distribution datasets, demonstrating strong generalization and resilience to population shifts. It is important to note that evaluating the predicted knowledge-informed metrics (as performed by \methodz) against standard state-of-the-art (SOTA) baselines is not directly applicable, as we are the first to propose and systematically evaluate the prediction of knowledge-informed metrics as an intermediate representation. Therefore, the evaluation of \methodz focuses on demonstrating how well it performs this novel task when compared to conventional baselines.

\begin{figure*}
    \centering
    \begin{subfigure}[b]{\textwidth}
        \centering
        \includegraphics[width=0.99\linewidth]{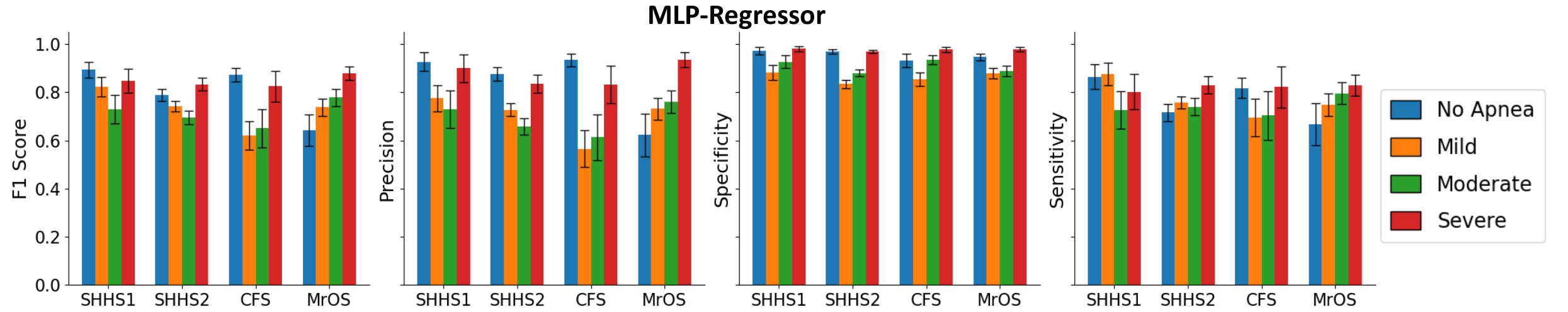}
        
        \captionsetup{labelformat=empty}
        \caption{(a) \method performance across various labels using \mlp. The figure illustrates higher accuracy in estimating patients with no apnea and severe apnea, with slightly lower performance for moderate and mild apnea predictions.}
    \end{subfigure}
    
    \vspace{1em}
     \begin{subfigure}[b]{\textwidth}
        \centering
        \includegraphics[width=0.99\linewidth]{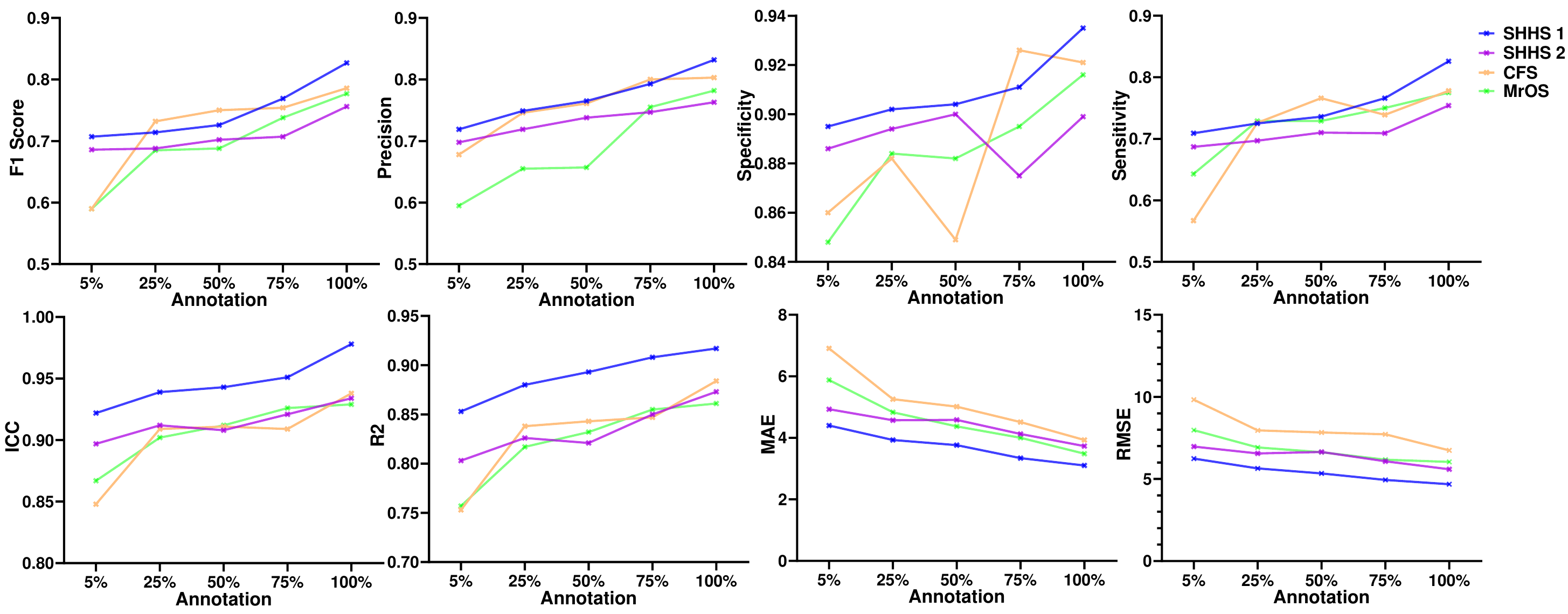}
        \captionsetup{labelformat=empty}
        \caption{(b) Comprehensive Comparison of Outcomes Using Varying Proportions of knowledge-informed metrics from 5\% to 100\% during the Experimental Setup. The results indicate an improvement across all metrics as the proportion of knowledge-informed metrics increases from 5\% to 100\%.}
    \end{subfigure}
     \caption{Outcome comparison across varying proportions of knowledge-informed metrics.}
    \label{fig:composite}
    \label{ap_1}
\end{figure*} 

\subsubsection{Regression \& Classification}
To assess the performance of \mlp model in the regression task of estimating the AHI, we utilize $R^2$ and $ICC$ as key metrics. As shown in Figures \ref{fi6}a and \ref{fi6}b, and Table \ref{1}, for \mlp, the $R^2$ and $ICC$ values were 0.917 and 0.957 for SHHS1, 0.873 and 0.934 for SHHS2, 0.884 and 0.938 for CFS, and 0.902 and 0.951 for MrOS. This indicates that the values estimated by the \method are most closely aligned with the true AHI values.

When the estimated AHI values are converted to their corresponding severity levels, as shown in Table \ref{1}, our proposed pipeline \method achieves good performance on all three datasets across $F_1$ Score, precision, sensitivity, and specificity, with confidence intervals set at 95\%. These results show that our model is robust and generalized to external datasets. Building on these observations, the confusion matrix in Figure \ref{fi6}c highlights \method's ability to perform better at identifying severe and healthy cases compared to mild and moderate ones in MrOS and CFS. In contrast, for SHHS1, \method demonstrates improved performance in identifying healthy, mild, and severe cases relative to moderate cases. Similarly, in SHHS2, \method performs particularly well in identifying severe cases compared to other severity levels. Figure \ref{ap_1}a, which presents the $F_1$ Score, precision, sensitivity, and specificity across various severity levels, further corroborates these findings.

\subsection{Effect of Knowledge-Informed Metrics}
To evaluate the role of knowledge-informed metrics in model performance, we conduct a series of experiments where we vary the quality and availability of these annotations during training. It is important to note that in our framework, knowledge-informed metrics are only used during training and are predicted from raw oximetry signals during inference. This reflects real-world deployment conditions, where sleep annotations are not accessible.

Figure~\ref{ap_1}b illustrates the effect of using 5\% to 100\% of correct knowledge-informed metrics during training. As the number of accurate annotations increases, we observe consistent improvements in downstream performance metrics, including $F_1$ Score, precision, sensitivity, specificity, ICC, and $R^2$—and reductions in $MAE$ and $RMSE$. This trend highlights that the model learns better feature representations when exposed to a greater quantity of clinically meaningful supervision. To further explore the importance of annotation quality, we simulate degraded annotations using two approaches: (1) randomly generated annotations and (2) shuffled segments of the ground-truth labels. Table~\ref{apx_6} shows a marked drop in performance under these conditions. This confirms that the model does not merely memorize annotations but instead relies on them to structure the learning of clinically relevant representations.

These findings support the premise that the knowledge-informed metrics serve as a valuable, interpretable supervision mechanism. The model benefits from their guidance only when they are accurate and derived from expert-like inputs, reinforcing the validity of using them as an intermediate clinical representation during training.

\begin{figure*} 
    \centering
    \includegraphics[width=0.98\linewidth]{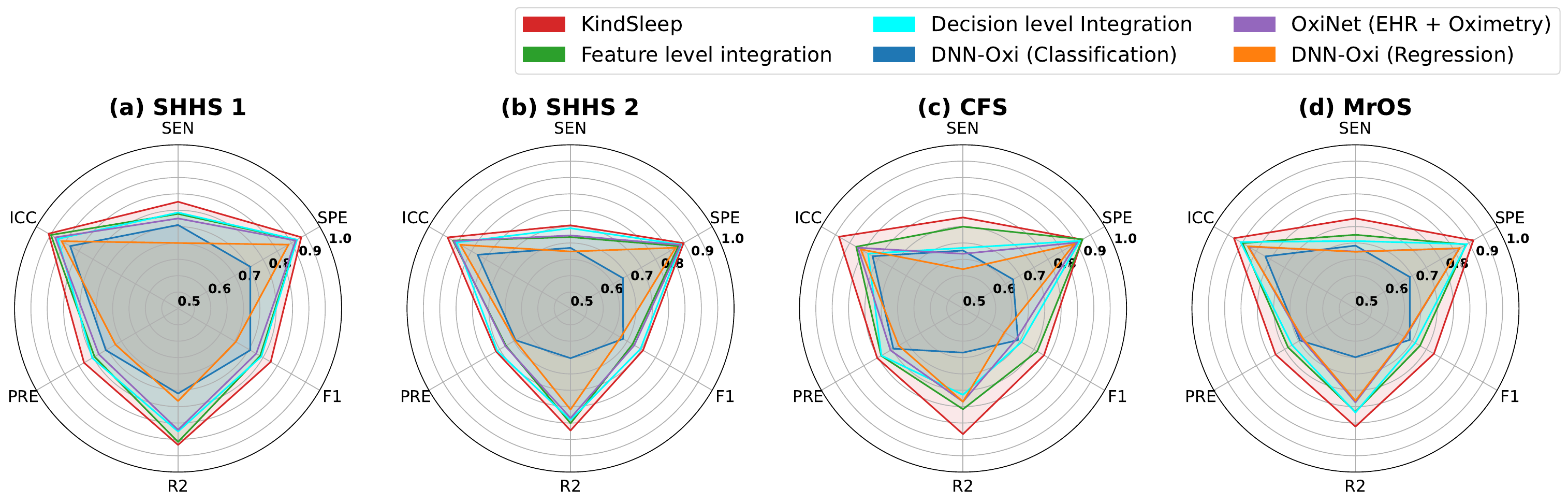}
    \caption{Radar charts comparing various performance metrics of our \method model against two baseline multimodal integration methods on the (a) SHHS1, (b) SHHS2, (c) CFS.}
    
    \label{fi8}
\end{figure*}

\begin{figure}
    \centering
    \begin{tikzpicture}
        \node (img1) at (0, 0) {\includegraphics[width=1\linewidth]{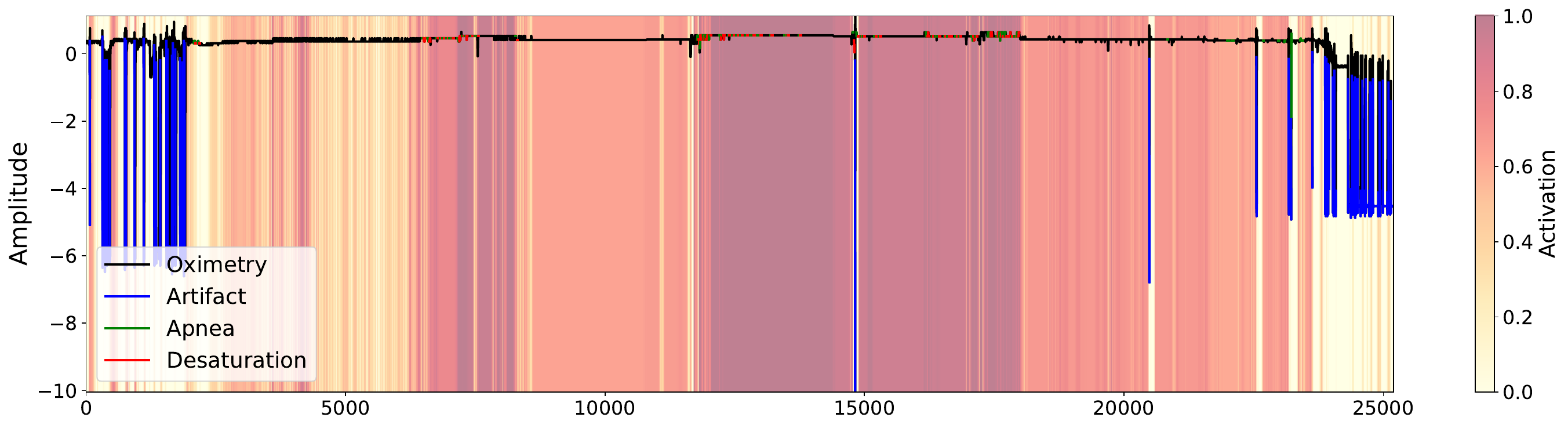}};
        \node (img2) at (0, -3) {\includegraphics[width=1\linewidth]{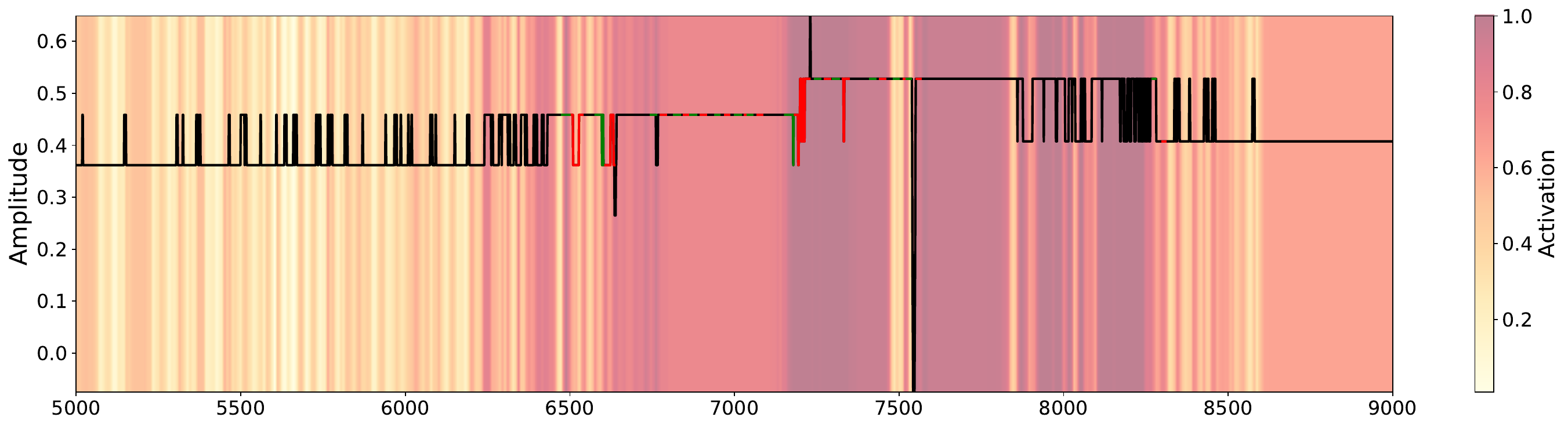}};
        \node (img3) at (0, -6) {\includegraphics[width=1\linewidth]{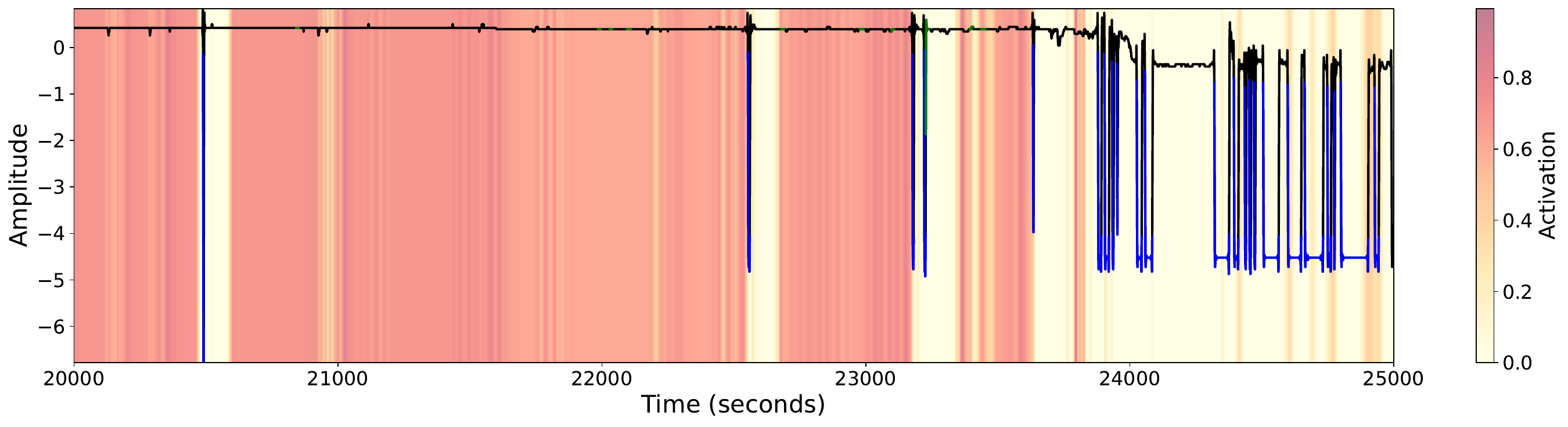}};
        \draw[thick, ->, black, line width=0.85mm] (img1.south) -- (img2.north);
        \draw[thick, ->, black, line width=0.85mm] (img2.south) -- (img3.north);   
    \end{tikzpicture}
    \caption{Attention mechanism employed by the \methodz model across oximetry signals, with events (e.g., desaturation, apnea, and artifacts) identified from ground truth annotations. The top section displays the global signal over the full duration (0–25,200 seconds), highlighting areas of high activation that correspond to physiologically relevant events, such as desaturation and apnea, while effectively ignoring artifact-prone regions. The middle section provides a focused view of the signal between 5,000 and 9,000 seconds, where the model demonstrates precise attention on desaturation events and hypopneas. The bottom section zooms into a segment from 20,000 to 25,000 seconds, showcasing the model’s ability to neglect low-amplitude, artifact-dominated regions devoid of clinically significant activities.}
    \label{fig_shap6}
    \vspace{-2ex}
\end{figure}

\begin{figure} 
    \centering
    \includegraphics[width=0.9\linewidth]{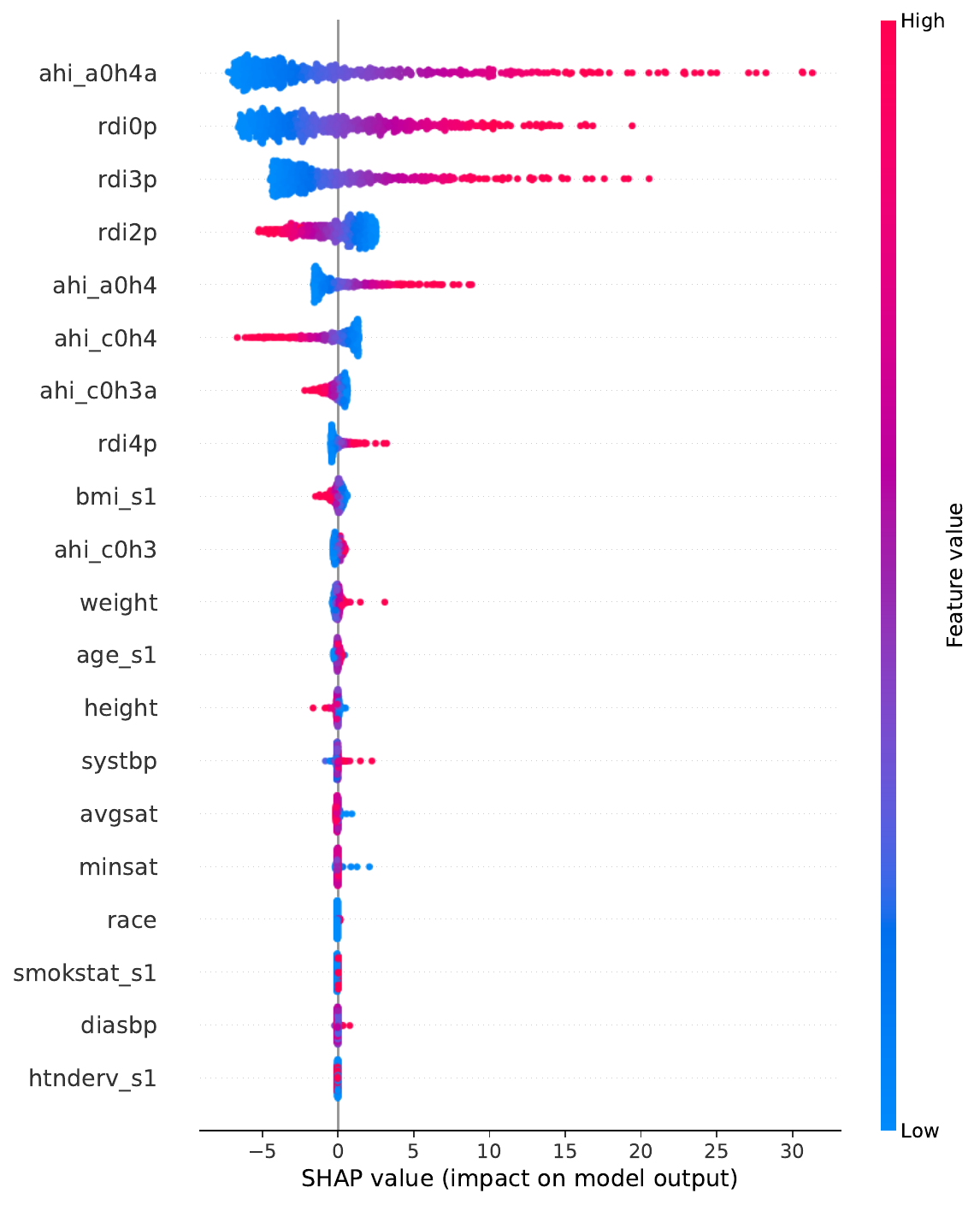}
    \caption{\mlp: SHAP results show the ranking of predicted knowledge-informed metrics against the patient clinical data.}
    \label{fig_shap1}
\end{figure}

\begin{figure}[h]
    \centering
   \begin{subfigure}[b]{0.485\linewidth}
        \centering
    \includegraphics[width=\linewidth]{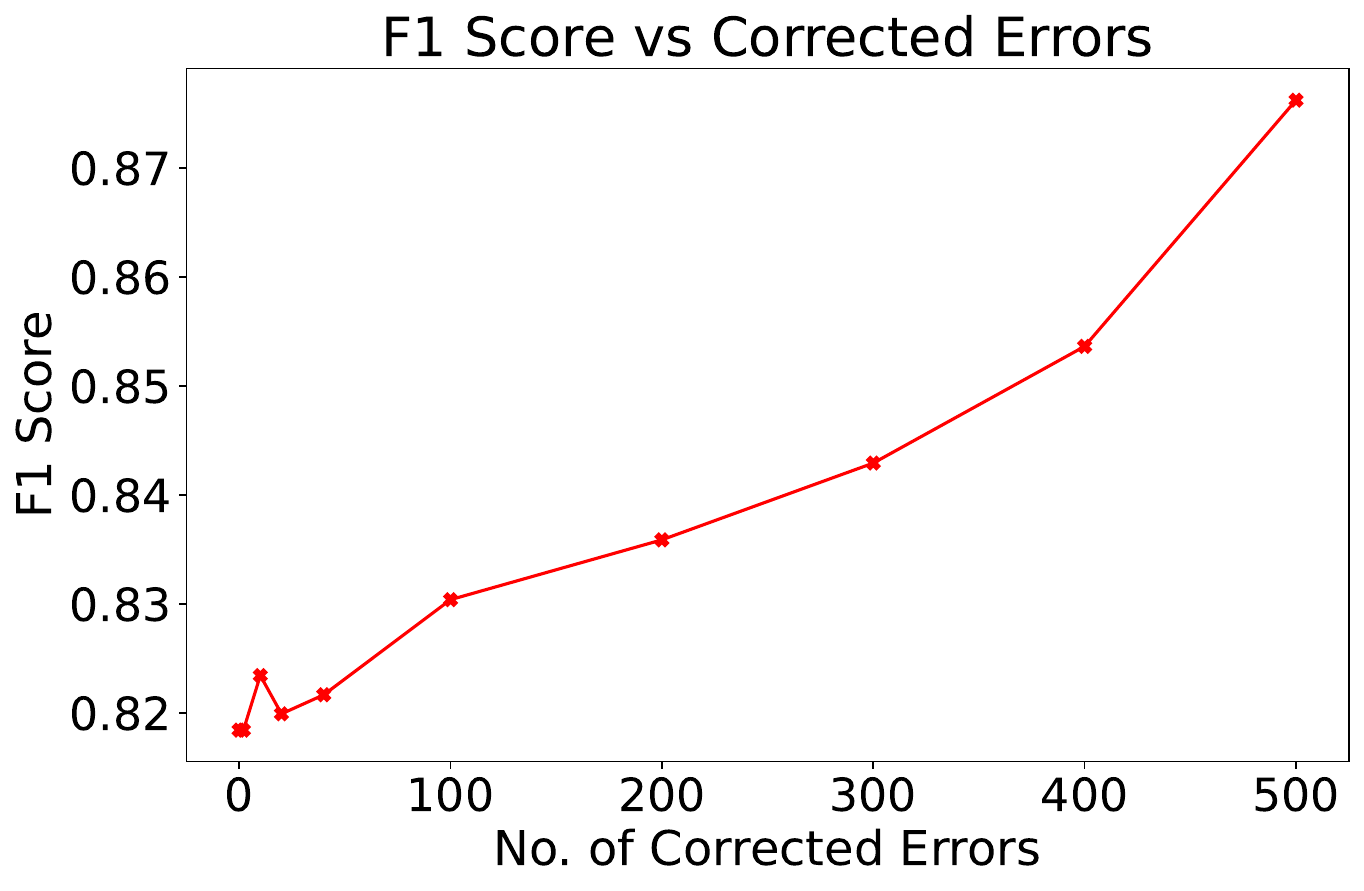}
    \end{subfigure}
    \begin{subfigure}[b]{0.485\linewidth}
        \centering
    \includegraphics[width=\linewidth]{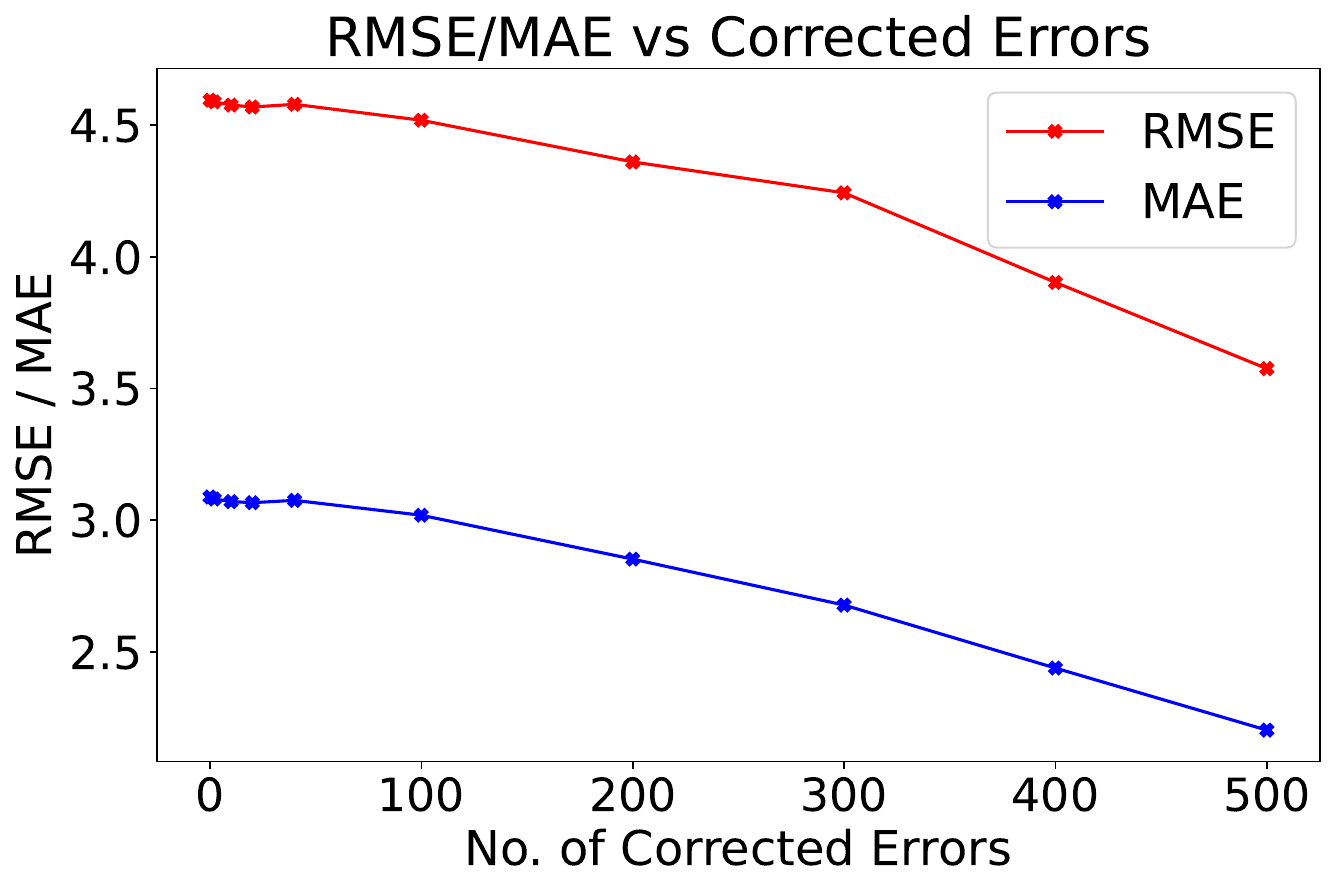}
    \end{subfigure}
      \caption{Relationship between the $F_1$ scores, MAE, and RMSE as errors are intercepted. We observed that identifying and adjusting these errors before passing them to the regression model during training significantly improves the system architecture's performance.}
    \label{fig_shap4}
\end{figure}

\begin{table*}
\centering
\caption{Impact of Incorrect, Partially Correct, and Fully Correct Annotations on Pipeline Performance. The results indicate that \method performance decreases with incorrect annotations.}
\label{apx_6}
\resizebox{\linewidth}{!}{
\begin{tabular}{ccccccccc}
\hline\hline
\textbf{Annotations} & \textbf{$F_1$ Score} & \textbf{Precision} & \textbf{Sensitivity} & \textbf{Specificity} & \textbf{$R^2$} & \textbf{ICC} &\ \textbf{MAE} &\ \textbf{RMSE}  \\ \hline
incorrect annotations & $0.223 \pm 0.036$     & $0.174 \pm 0.032$   & $0.313 \pm 0.038$  & $0.749 \pm 0.020$ & 0.125 & 0.245 & 10.831 & 15.226\\
incorrect + 5\% correct annotations & $0.637 \pm 0.039$     & $0.647 \pm 0.040$    & $0.643 \pm 0.040$      & $0.871 \pm 0.017$ & 0.832 & 0.910 & 4.854 & 6.679\\ 
correct annotations & $0.827 \pm 0.030$    & $0.832 \pm 0.030$    & $0.826 \pm 0.034$ & $0.935 \pm 0.013$ & 0.917 & 0.957 & 3.101 & 4.677\\

\hline \hline
\label{apx_1}
\end{tabular}}
\end{table*}

\subsection{Model Comparison}
For performance comparison, we have considered two existing multimodal combination methods \cite{venugopalan2021multimodal} using the same model parameters to show the advantage of integrating the knowledge-informed metrics:
\begin{itemize}[leftmargin=0.6cm]
\item {\verb|Feature level integration|}: High-level features extracted from each modality are concatenated before passing into a model for decision-making.
\item {\verb|Decision level integration|}: Voting is performed using decisions of individual modality model. 
\end{itemize}

We have also considered three SOTA baselines for OSA diagnosis \cite{levy2023deep, chi2024apnea,chen2023deep}:
\begin{itemize}[leftmargin=0.6cm]
    \item {\verb|DNN-OXi (Classification)|\cite{chen2023deep}}: Defined the solution to the problem as a classification problem using a customized DNN model and an oximetry signal.
    \item \verb|DNN-Oxi (Regression)| \cite{chi2024apnea}: A deep neural network trained on raw oximetry signals, formulated as a regression task to estimate continuous AHI values.
    \item {\verb|OxiNet (Multimodal)|\cite{levy2023deep}}: Introduced an OxiNet model to combine clinical data and oximetry signal to estimate and classify patient's AHI.
\end{itemize}

It is important to note that both our method and the referenced baseline models process the full oximetry signal without partitioning into smaller chunks or windows. We deliberately selected baseline models that followed this same assumption to ensure a fair and consistent comparison. While input processing strategies can vary in the literature, our comparisons are grounded in models that align with our signal-level granularity.
As shown in Figure \ref{fi8}, we observe that \method outperforms existing literature and data integration methods for multimodal fusion. This indicates that knowledge-informed metrics are more significant and have a greater impact on model decisions when compared to existing literature.

\section{Discussion}
While deep learning models have the potential to learn and predict AHI independently without human intervention~\cite{wu2024clinical,wu2024developing}, \method differs in that its learning relies entirely on knowledge-informed metrics and control to perform this task. This dependency makes \method more efficient, robust and provides a clear direction during the learning process. Although we have demonstrated that \method significantly outperforms previous integration methods and existing literature, we now focus on discussing the transparency, trust, and interpretability aspects of this pipeline.

To begin, we investigate how effectively \methodz estimates knowledge-informed metrics. For this purpose, we analyze oximetry signals using an attention mechanism derived from our trained model. Specifically, we modified the original Gradient-weighted Class Activation Mapping (Grad-CAM) technique \cite{selvaraju2017grad} to align with the requirements of our study.  This adaptation enables us to visualize neural attention through an attention map, as shown in Figure~\ref{fig_shap6}. These maps highlight the model’s focus on various regions of the signal, effectively revealing its attention to clinically meaningful patterns. Closer examination shows that the model directs its attention primarily toward segments of the oximetry signal corresponding to desaturation and apnea events. These high-activation areas align with significant drops in oxygen saturation, reinforcing that the model is focusing on physiologically relevant regions rather than stable signal segments. While it may initially appear that some activations lie in stable regions, further inspection reveals that these often precede or lead into desaturation transitions, indicating early detection. Importantly, regions dominated by artifacts are largely ignored, demonstrating the model's robustness in distinguishing signal noise from clinical indicators.

Secondly, we analyze how independent features influence the regression model using the \mlp. To achieve this, we utilize SHapley Additive exPlanations (SHAP) \cite{lundberg2017unified}, which quantify the contribution of each feature (clinical data and knowledge-informed metrics) to the final prediction. This enhances our understanding of which input factors most significantly influence the model's estimation and classification of patient AHI. As depicted in Figure~\ref{fig_shap1}, the estimated knowledge-informed metrics have a more substantial influence on model predictions compared to the demographic features. Specifically, annotations such as ahi\_a0h4a, ahi\_a0h4, ahi\_c0h3a, ahi\_coh3, rdi0p, and rdi3p emerge as the most influential features. These metrics integrate clinically interpretable constructs like desaturation levels, minimum saturation, and hypopnea frequency. For example, ahi\_a0h4a quantifies the number of apneas and hypopneas with $\geq$ 4\% oxygen desaturation or with arousal per hour of sleep. rdi0p includes apneas and hypopneas with no desaturation threshold, with or without arousal, normalized by sleep time. In contrast, \texttt{rdi3p} applies a 3\% desaturation threshold, adding more specificity. These annotations align closely with clinical guidelines for evaluating sleep-related breathing disorders.

Furthermore, among all clinical features, BMI emerges as a co-contributing factor. Analysis of the relationship between BMI and AHI reveals a trend: individuals with higher BMI values tend to have more severe AHI outcomes. Stratification of SHHS1 test dataset participants into BMI categories demonstrates that obese individuals exhibit the highest mean AHI ($21.821\pm 19.302$), followed by overweight individuals ($15.688 \pm 14.120$) and normal-weight individuals ($11.716 \pm 13.855$). These findings suggest that BMI plays a role in influencing the severity of sleep-related breathing disorders. Correlation analysis further supports this relationship, with Pearson ($r$ = \( 0.306 \), \( p_{value} < 10^{-13} \) and Spearman ($\rho$ = $0.305$, \( p_{value} < 10^{-13} \)) coefficients indicating weak to moderate but statistically significant positive associations between BMI and AHI. These results reinforce the hypothesis that increased body mass may contribute to airway obstruction and other physiological changes that exacerbate sleep apnea symptoms \cite{isono2009obstructive, qureshi2003obstructive, eckert2008pathophysiology}. In terms of predictive performance, the $F_1$ scores highlight BMI's role in distinguishing severity across different categories. For obese individuals, the $F_1$ score was \( 0.818 \pm 0.059 \), while overweight individuals achieved an $F_1$ score of \( 0.825 \pm 0.047 \). Among normal-weight individuals, the $F_1$ score was slightly higher at \( 0.839 \pm 0.053 \). All $F_1$ scores are reported with 95\% confidence intervals, demonstrating consistent performance across BMI categories.

Another advantage of \method compared to the existing integration pipeline is that it allows for model improvement through human intervention since the model learns from sleep annotation and supervision. Figure \ref{fig_shap4} shows that since clinicians can understand the guided model decisions, identifying errors that might occur at training time during \methodz estimation of the knowledge-informed metrics will significantly improve the model's performance. One drawback to this is that the intervention only influences the performance of the regression model and does not affect the ground truth of the \methodz, therefore someone needs to keep intervening. To address this limitation, we will consider continuous learning, where \methodz is designed to continuously learn as new oximetry data, collected from different calibrated devices, are fed into the system. This can be built by adopting the independent or sequential learning schema introduced in concept-based learning through a single modality \cite{koh2020concept, nnamdi2023concept}.

{Beyond this technical direction, several broader considerations remain. First, while interpretability was demonstrated qualitatively, quantitative analyses, such as correlations between attention weights and oxygen desaturation events or consistency of SHAP-derived importance scores, will be needed to confirm alignment with clinically meaningful features. Second, calibration analyses using reliability diagrams and Brier scores need to be incorporated to ensure that \method’s probability estimates are trustworthy in clinical contexts. Third, because the model currently emphasizes oxygen desaturation, future extensions may benefit from integrating additional low-cost signals (e.g., respiratory effort or airflow surrogates) and waveform morphology features to better capture hypopnea events.}
\section{Conclusion}
\label{sec:conclusion}
In this paper, we have proposed a novel pipeline, \method, which consists of \methodz and a regression model for estimating and classifying a patient's AHI. The proposed pipeline relies on knowledge-informed metrics to guide its learning, making it better than existing multimodal integration pipelines that solely depend on features extracted based on patterns observed by the model. The robustness and generalizability of \method are demonstrated through evaluations on multiple datasets. Additionally, we have shown the transparency of the pipeline, which is essential for practicing responsible AI and potential adoption by physicians. This approach aligns with the principles of CBM, further demonstrating the utility of knowledge-informed metrics in enhancing model interpretability and performance.
For future work, we will incorporate continuous learning to enable the model to continuously learn from its mistakes and correct itself, thereby preventing repetition of errors and improving its performance over time.

% \vspace{-1mm}
\section*{Acknowledgment}
This research was supported by a seed research grant from Shriners Children’s Hospital. Additional support was provided in part by the AI Makerspace of the College of Engineering and other research cyberinfrastructure resources and services offered by the Partnership for an Advanced Computing Environment (PACE) at the Georgia Institute of Technology, Atlanta, Georgia, USA. We also gratefully acknowledge Wallace H. Coulter Distinguished Faculty Fellowship, a Petit Institute Faculty Fellowship, and research funding from Amazon and Microsoft Research to Professor May D. Wang.

\section*{Data Availability}
{This work used the Sleep Heart Health Study (SHHS) dataset. The SHHS was supported by National Heart, Lung, and Blood Institute cooperative agreements U01HL53916 (University of California, Davis), U01HL53931 (New York University), U01HL53934 (University of Minnesota), U01HL53937 and U01HL64360 (Johns Hopkins University), U01HL53938 (University of Arizona), U01HL53940 (University of Washington), U01HL53941 (Boston University), and U01HL63463 (Case Western Reserve University). The National Sleep Research Resource was supported by the National Heart, Lung, and Blood Institute (R24 HL114473, 75N92019R002). This work also made use of the MrOS Sleep Study dataset (Outcomes of Sleep Disorders in Older Men). The study was funded by the National Heart, Lung, and Blood Institute under grants R01 HL071194, R01 HL070848, R01 HL070847, R01 HL070842, R01 HL070841, R01 HL070837, R01 HL070838, and R01 HL070839. The National Sleep Research Resource was supported by the National Heart, Lung, and Blood Institute (R24 HL114473, 75N92019R002). This work further made use of the Cleveland Family Study (CFS) dataset. The study received support from the National Institutes of Health under grants HL46380, M01 RR00080-39, T32-HL07567, RO1-46380. The National Sleep Research Resource was supported by the National Heart, Lung, and Blood Institute (R24 HL114473, 75N92019R002).}

\bibliographystyle{ACM-Reference-Format}
\bibliography{ref}

\appendix
\clearpage

% \section{\textbf{Appendix}}
\newpage
\title{Supplementary}
\section{Data Preprocessing Details}
\label{sec:data}
\subsection{Clinical Data}
The clinical features utilized in this study included key demographic, anthropometric, cardiovascular, lifestyle, and comorbidity features directly extracted from patient records. Specifically, the following features were considered: ethnicity, body mass index (BMI), systolic blood pressure (SBP), diastolic blood pressure (DBP), age, current smoking status, race, weight (wtkg), height (htcm), and hypertension status (htnx). These features were chosen for their relevance to the clinical context of the study and their potential to influence the predictive performance of the models.
Minimal preprocessing was applied to the extracted clinical features to maintain the integrity and interpretability of the raw clinical information. Instances with missing data in any of the features were excluded to maintain the consistency and quality of the dataset. This ensured that the model only learned from complete and reliable data, reducing potential noise or bias from imputation.
Categorical variables were encoded into numerical formats for direct use in the modeling framework. One-hot encoding was applied to create binary vector representations for multi-class categories, while label encoding was used for binary variables. These encoding strategies ensured that categorical variables contributed effectively to the predictive model without introducing artificial ordinal biases. The resulting features were represented as vectors without further transformations or dimensionality reduction. These vectors were concatenated with intermediate predictions from \methodz, creating a unified feature set. This integration preserved the granularity of the clinical data while enhancing the feature space with learned representations from the predictive model. The resulting feature set was used as input to a regression model.

\begin{table*}[]
\centering
\caption{Overview of dataset details, including training, validation, and testing splits, participant counts, demographics (age, BMI, gender, ethnicity, and race), and AHI categories (Healthy, Mild, Moderate, Severe) with corresponding mean and standard deviation.}
\begin{tabular}{lccccccccc}
\toprule
\textbf{Description} & \textbf{Train}  & \textbf{Validation} & \textbf{Test}\\ \midrule
Recording (Counts) & 3539 & 1531 & 4745 \\
Duration (Hours) & 24,773 & 10,717 & 33,215\\
Gender(Male/Female) & 47.40/52.67 & 48.07/51.93 & 57.74/42.26\\
Age (Years) & $63.27 \pm 11.23$ & $63.47 \pm 10.87$ & $65.57 \pm 15.42$\\
Body Mass Index (BMI) & $28.22 \pm 5.10$ & $27.99 \pm 4.97$ & $28.46 \pm 5.66$\\
Ethnicity (Non-Hispanic/Hispanic) & 95.51/4.49 & 94.91/5.09 & 97.01/2.99\\
Race (White/Black/Others) & 85.33/8.62/6.05 & 45.36/51.18/3.46 & 82.11/12.46/5.44\\ 
\hline
\textbf{Apnea-Hypopnea Index (AHI)}\\
\hline
Healthy & $2.45 \pm 1.40$ & $2.34 \pm 1.40$ &  $2.32 \pm 1.39$ \\
Mild & $9.30 \pm 2.81$ & $9.27 \pm 2.82$ & $9.35 \pm 2.81$\\
Moderate & $20.92 \pm 4.11$ & $21.08 \pm 4.41$ & $21.15 \pm 4.22$\\
Severe & $47.69 \pm 16.78$ & $46.37 \pm 16.64$ & $46.12 \pm 14.98$\\
\bottomrule 
\end{tabular}
\label{xd}
\end{table*}

\subsection{Oximetry Signal}
To ensure signal consistency, reduce noise, and prepare the data for training, we employed a comprehensive preprocessing pipeline. Proper preprocessing is essential for mitigating uncertainties and artifacts that can significantly impact analysis outcomes and degrade signal quality. The steps in this pipeline include:
\begin{itemize}[leftmargin=0.6cm]
    \item Signals shorter than a predefined length of 25,200 data points were padded with zeros, while signals exceeding this length were truncated. This padding or truncation ensured that all signals had consistent input dimensions.
    \item High-frequency noise is a common issue in oximetry signals, as it can distort the data and compromise analysis accuracy. To address this issue, we employed the Savitzky-Golay (savgol) filter \cite{savitzky1964smoothing}, a low-pass filter widely recognized for its capability to smooth signals while preserving the underlying signal structure and trends. The savgol filter operates by fitting successive subsets of the signal to low-degree polynomials, thereby minimizing noise without distorting the inherent structure of the data. In alignment with the literature, savgol filter was selected for its proven efficacy in noise reduction and artifact suppression, making it particularly suitable for this application \cite{lima2022oxitidy, contreras2024estimation, chakraborty2024non}.
    \item Missing or non-physiological values within the signals were addressed using linear interpolation. This approach preserved the continuity of the signal and mitigated the potential impact of data loss.
    \item Lastly, all signals were normalized to ensure consistency. Standardization was performed by transforming each signal to have a mean of zero and a standard deviation of one, a process that removes scale differences and centers the data around zero. 
\end{itemize}
Unlike conventional approaches that partition signals into smaller segments (e.g., 30-second windows), the entire preprocessed signal of 25,200 data points was retained and utilized as a single sequence for training. This approach was designed to preserve the temporal dependencies and long-term patterns within the data, enabling the model to capture clinically relevant trends more effectively \cite{kang2024introducing}.

\section{SLeep Annotation Model (\methodz) Details}
\label{sec:method1}
\methodz, which relies on human control and supervision, incorporates a deep attention layer (DAL) to improve the estimation and analysis of time series data (oximetry signal). The model takes an input sequence $\mathbf{X} \in \mathbb{R}^{n \times d}$ and a set of parameters $\theta$ to generate concepts $\mathbf{C} \in \mathbb{R}^{m}$. First, the input is reshaped to the dimensions $(n, d)$. Then, the reshaped input passes through a series of $N_c$ convolutional layers, each consisting of $f_i$ filters with kernel size $k_i$, followed by batch normalization, leaky rectified linear unit (ReLU) activation with parameter $\alpha$, and max pooling with pool size $p$ and stride $s$. The output of the convolutional layers is then processed by two Bidirectional LSTM (BiLSTM) layers, each with a hidden state dimension of $h$. To avoid overfitting, dropout regularization with rate $r$ is applied to the BiLSTM output.

Regarding the BiLSTM processing, \methodz extends the traditional LSTM framework by processing the data in both forward and reverse directions, capturing dependencies that occur at different time scales as follows:
\begin{align}
\overrightarrow{\mathbf{h}_t} &= \text{LSTM}(\mathbf{x}_t, \overrightarrow{\mathbf{h}_{t-1}}; \theta_{\text{fw}}), \\
\overleftarrow{\mathbf{h}_t} &= \text{LSTM}(\mathbf{x}_t, \overleftarrow{\mathbf{h}_{t+1}}; \theta_{\text{bw}}),
\end{align}
where $\overrightarrow{\mathbf{h}_t}$ and $\overleftarrow{\mathbf{h}_t}$ represent the hidden states at time $t$ for the forward and backward LSTMs, respectively. The outputs from both directions are combined to form a unified representation:
\begin{equation}
\mathbf{h}_t = [\overrightarrow{\mathbf{h}_t}; \overleftarrow{\mathbf{h}_t}],
\end{equation}
This combined output $\mathbf{h}_t$ captures information from both past and future contexts, leading to a more comprehensive understanding of the input sequence.
% \vspace{-2.5mm}

Following the BiLSTM and dropout layer, the model applies DAL (\ref{sec:method3}) to weigh the importance of each timestep's output. A Dense layer is then applied to the output from DAL to estimate the knowledge-guided metrics. Training of the model involves compiling with an Adam optimizer and using a mean absolute error loss function:
\begin{equation}
\mathcal{L} = \frac{1}{n} \sum_{i=1}^{n} |\mathbf{C}_i - \hat{\mathbf{C}}_i|,
\end{equation}
where $\hat{\mathbf{C}}_i$ denotes the model's prediction of knowledge-guided metrics for the $i$-th sample.
% \vspace{-3mm}

\methodz learns knowledge-guided metrics through a combination of convolutional layers, BiLSTM layers, and DAL. The convolutional layers capture local patterns and features from the input signals, while the BiLSTM layers model the temporal dependencies. The DAL allows the model to focus on the most relevant parts of the input sequence to make predictions. By combining these components, our model can effectively learn and estimate knowledge-guided metrics from raw input signals.
\section{Deep Attention Layer (DAL) Details}
\label{sec:method3}

\methodz integrates DAL to focus on the most relevant parts of the input sequence. DAL takes an input sequence $\mathbf{x} = (x_1, \ldots, x_T)$ and a set of attention units $u$ to generate a context vector $\mathbf{c}$. The layer first initializes the attention weight matrix $\mathbf{W} \in \mathbb{R}^{d \times u}$, context weight matrix $\mathbf{W}_c \in \mathbb{R}^{u \times 1}$, and attention bias $\mathbf{b} \in \mathbb{R}^{T \times u}$~\cite{wan2023swift}. It then computes the attention features $\mathbf{F}$ using:
\begin{equation}
\mathbf{F} = \tanh(\mathbf{W} \mathbf{x} + \mathbf{b}).
\end{equation}
Next, the attention scores $\mathbf{e}$ are calculated using:
\begin{equation}
\mathbf{e} = \mathbf{F} \mathbf{W}_c,
\end{equation}
\begin{equation}
\mathbf{e} = \textit{squeeze}(\mathbf{e}, \text{axis}=-1),
\end{equation}
followed by a squeeze operation on the last axis. The attention weights $\boldsymbol{\alpha}$ are then computed by applying a softmax function to $\mathbf{e}$:
\begin{equation}
\boldsymbol{\alpha} = \textit{softmax}(\mathbf{e}).
\end{equation}
If $\boldsymbol{\alpha}$ is not expanded, an expand dims operation is applied on the last axis. Finally, the context vector $\mathbf{c}$ is initialized as a zero vector and updated iteratively for each time step $t$ using:
\begin{equation}
\mathbf{c} = \mathbf{c} + \mathbf{x}_t \odot \boldsymbol{\alpha}_t,
\end{equation}
where $\odot$ denotes element-wise multiplication.The context vector $\mathbf{c}$ is then returned as the output of the DAL.

\section{Regression Model Details}
\label{sec:method2}
The regression models considered in this study for estimating the AHI after concatenating the estimated sleep annotation with the vectors from the clinical data is the Multilayer Perceptron Regressor (\mlp). This model have gained significant traction in the realm of machine learning for their ability to model complex nonlinear relationships. Unlike traditional linear regression models, \mlp model leverages a feed-forward mechanism to capture intricate patterns within data, making them suitable for a wide range of regression tasks. \mlp models are among the numerous neural network designs that are basic in framework, simple to execute, and have strong fault tolerance, resilience, scalability, and outstanding nonlinear mapping capabilities \cite{wu2022hybrid, shams2024water}.

\section{Evaluation Metrics} The following evaluation metrics were considered for evaluating the regression task (before converting the estimated AHI into the four levels of severity),
\begin{itemize}
    \item \textbf{Coefficient of Determination}, R-squared ($R^2$) measures how well the independent variable(s) in a statistical model explains the variation in the dependent variable.
\begin{equation}
    R^2 = 1 - \frac{\sum_{k=1}^{n} (y_k - \hat{y}_k)^2}{\sum_{i=1}^{n} (y_k - \bar{y})^2},
\end{equation}
where $y_k$ are the observed values, $\hat{y}_k$ are the predicted values, and $\bar{y}$ is the mean of the observed values.
\item \textbf{Intraclass Correlation Coefficient} ($ICC$) measures the reliability of estimated AHI. $ICC$ is subject to a variety of statistical assumptions such as normality and stable variance, which are rarely considered in health applications \cite{bobak2018estimation}. Mathematically, it is expressed as

\begin{equation}
    ICC = \frac{MS_1 - MS_w}{MS_1 + (k - 1) MS_w + \frac{k}{n}(MS_i - MS_w)},
\end{equation}
where $MS_1$ is the instance mean square, $MS_w$ is the mean square error, $MS_i$ is the observers mean square and $k$ is the number of observation.
\end{itemize}

After the conversion of the estimated AHI to the four levels of severity, the confusion matrix, precision, $F_1$ score, sensitivity, and specificity were used to evaluate the accuracy of the model's estimated values.

\end{document}